\documentclass[11pt,a4paper]{article}
\usepackage{times,latexsym}
\usepackage{url}
\usepackage[T1]{fontenc}

\usepackage[]{acl}

\usepackage{xspace,mfirstuc,tabulary}

\newif\iftaclinstructions
\taclinstructionstrue 
\iftaclinstructions
\renewcommand{\confidential}{}
\renewcommand{\anonsubtext}{(No author info supplied here, for consistency with
TACL-submission anonymization requirements)}
\newcommand{\instr}
\fi
\taclpubformattrue
\iftaclpubformat 

\else

\fi

\usepackage[utf8]{inputenc}
\usepackage{pdflscape}
\usepackage{multirow}
\usepackage{makecell}
\usepackage{booktabs}
\usepackage{rotating}
\usepackage{amsmath}
\usepackage{subfigure}
\usepackage{colortbl}
\usepackage{algorithmicx,algorithm}
\usepackage{amsfonts}

\usepackage{microtype}

\title{\textsc{AdaVAE}: Exploring Adaptive GPT-2s in Variational Auto-Encoders for Language Modeling}

\author{Haoqin Tu,
  {Zhongliang Yang},
  Jinshuai Yang,
  {Yongfeng Huang}\\
  Department of Electronic Engineering \\
  Tsinghua University, Beijing, 100084, China\\
  \texttt{tuisaac163@gmail.com}, \texttt{yangzl15@tsinghua.org.cn}\\ \texttt{yjs20@mails.tsinghua.edu.cn}, \texttt{yfhuang@tsinghua.edu.cn}}

\begin{document}
\maketitle

\begin{abstract}
Variational Auto-Encoder (VAE) has become the de-facto learning paradigm in achieving representation learning and generation for natural language at the same time. Nevertheless, existing VAE-based language models either employ elementary \textsc{RNN}s, which is not powerful to handle complex works in the multi-task situation, or fine-tunes two pre-trained language models (PLMs) for any downstream task, which is a huge drain on resources. In this paper, we propose the first VAE framework empowered with adaptive GPT-2s (\textsc{AdaVAE}). Different from existing systems, we unify both the encoder\&decoder of the VAE model using GPT-2s with adaptive parameter-efficient components, and further introduce \textit{Latent Attention} operation to better construct latent space from transformer models. Experiments from multiple dimensions validate that \textsc{AdaVAE} is competent to effectively organize language in three related tasks even with less than $15\%$ activated parameters in training. Our code is available at \url{https://github.com/ImKeTT/AdaVAE}.
\end{abstract}

\section{Introduction}
With the development of natural language processing (NLP) techniques, neural networks have been introduced to handle various tasks and empirically promoted their performances to higher levels. As a competitive solution to miscellaneous NLP tasks, the variational auto-encoder (VAE) model \cite{bowman2015generating} is not only a powerful generative model but a textual feature learning framework when trained properly. Its structured and continuous hidden space makes it easy to derive high-level linguistic knowledge \cite{fang2019implicit} for either generation or understanding. However, some problems in practice may limit the modeling capacity and empirical performance of VAE-based language models. One of the major challenges that text VAE faces is its weak latent representation issue, which may induce other related problems in the model, such as KL collapse problem \cite{bowman2015generating}, latent vacancy issue \cite{xu2020variational} and token-latent inconsistency \cite{shen2020educating}. Several approaches in both modeling architecture and training schedules have been devised to handle these issues \cite{zhao2017infovae,fu2019cyclical,zhao2017learning}. These methods share a similar goal to enhance the expression of the VAE encoder for constructing meaningful latent space $\mathcal{Z}$ to make it compatible with their decoders. 

Lately, as pre-trained language models (PLMs) are becoming the cornerstone of many state-of-the-art methods in NLP tasks, their potential has been widely explored. An intuitive idea comes to mind to enhance learned latent representations in VAE is that: using two well-matched PLM encoders and decoders to VAEs, so their latent spaces can be both easily derived from the training data and infused into the generation process. Recent works have sought to incorporate large-scale PLMs such as \textsc{BERT} \cite{devlin2018bert} and \textsc{GPT-2} \cite{radford2019language} into VAE models, which strengths VAEs in various tasks, including natural language generation (NLG) and understanding (NLU) \cite{li2020optimus,park2021finetuning,fang2021transformer}. While these ``big VAEs'' promote model performance to a higher level, they also bring much larger parameters to be trained compared with \textsc{RNN}-based ones or a single PLM. For instance, \textsc{Optimus} \cite{li2020optimus} need to tune at least one \textsc{BERT} and one \textsc{GPT-2} model (not counting middle layers) with over 220 million parameters for any downstream task, which is intolerant with the increase of workload.

On the other hand, introducing PLMs to VAE models strengthens their representation learning abilities, but also brings reflections on how to properly construct and utilize the latent space in PLMs. A VAE model cannot perform its best self without the proper latent guidance even with a pair of very powerful encoder-decoder. One direction to remit VAE's weak latent representation lies in improving the latent construction and infusion methods. Since there may remain discrepancies in the representation aspect between \textsc{RNN}-based and transformer-based VAEs, rethinking the latent knowledge generation and infusion method for ``big VAEs'' is vital for the fulfillment of their best potential.

To sum up, two shortcomings of a large-scale VAE can be stated as (1) \textbf{Excessive training parameters}: existing ``big VAEs'' fine-tune all the parameters in encoder\&decoder, which means at least two separate PLMs are fully activated during training. This leads to prohibitive computational overhead and low training efficiency of models in situations like multi-task learning. (2) \textbf{Incompatible latent spaces} with PLMs: most VAE models construct the latent space $\mathcal{Z}$ from the encoder through the last hidden state from its encoder and infusing it to the decoder by either initializing the decoder with latent vector or adding it to decoder hidden states \cite{bowman2015generating}. These methods may be suitable for exploiting undergrad textual features from \textsc{RNN} models but may be the otherwise for transformers. 

To address these problems, we propose \textsc{AdaVAE}. \textsc{AdaVAE} essentially comprises two adaptive parameter-efficient GPT-2s, which leverage powerful PLMs without excessive resource consumption. In detail, we add additional adapter components between feedforward layers and the output of an attention block to these GPT-2. For $\mathcal{Z}$ construction and infusion method, we first propose \textit{Latent Attention} that produces textual knowledge by considering the encoder's representations into the attention operation. Then we further investigate two existing latent knowledge fusion methods based on transformer models during the generation process. We further conduct studies to explore the effectiveness of proposed adapter components and latent space construction methods. Extensive experiments on several downstream tasks span six datasets, including language modeling, low resource classification, and guided text generation produce promising results w.r.t. model efficiency and automatic metrics. 

\noindent \textbf{Contributions.} (1) We propose an adapter module for parameter-efficient GPT-2 as both the encoder and decoder of \textsc{AdaVAE}. To our best knowledge, \textsc{AdaVAE} is the first ``big VAE'' model with unified parameter-efficient PLMs that can be optimized with minimum trainable parameters. (2) We devise \textit{Latent Attention} operation for latent space construction in \textsc{AdaVAE}. Varied parameter-efficient components and two latent knowledge infusion methods are further explored in our VAE model. (3) \textsc{AdaVAE} achieves state-of-the-art performance in language modeling and comparable performance in classification and controllable generation tasks respectively with only $14.66\%$ parameter activated.


\section{Related Work}
\subsection{Latent Variable Language Models}
VAE is famous for its continuous latent space, its extensions in the language domain have inspired new applications by exploiting many interesting properties of the model’s latent space. As a preliminary, the evidence lower bound (\textbf{ELBO}) of a VAE is:
\begin{equation}
\begin{aligned}
\underbrace{\mathbb{E}_{q(\boldsymbol{z} \mid \boldsymbol{X})}\left[\log p(\boldsymbol{X} \mid \boldsymbol{z}\right)]}_{\mathcal{L}_{\text{rec}}}-\underbrace{\mathbb{D}_{\mathrm{KL}}\left(q(\boldsymbol{z} \mid \boldsymbol{X}) \| p(\boldsymbol{z})\right)}_{\mathcal{L}_{\text{KL}}},
\end{aligned}
\label{eq:elbo}
\end{equation}
where $\boldsymbol{X}$ is the texts to be modeled, and $\boldsymbol{z}$ is the latent variable sampled from latent space $\mathcal{Z}$. 

In real world scenario, some defects limit the empirical performance of VAEs for language modeling including KL collapse issue \cite{bowman2015generating}, latent vacancy issue \cite{xu2020variational} and token-latent inconsistency problem \cite{shen2020educating}. Many related theories and solutions to these drawbacks were proposed including optimizing decoder architectures \cite{semeniuta2017hybrid,li2020optimus}, inventing auxiliary objectives \cite{xiao2018dirichlet,fang2019implicit,dai2020apo}, novel encoder training schedule \cite{bowman2015generating,fu2019cyclical}, incorporating flexible latent code posterior \cite{wang2019topic}, etc. These methods generally share the same goal: to impair the ability of a powerful decoder and strengthen the expression of latent space by reinforcing encoder ability. 

With more powerful transformer-based language models arising, researchers start to march on combining language VAEs with transformers \cite{vaswani2017attention}. For example, transformers are recently considered in VAEs for classification \cite{gururangan2019variational} and storytelling \cite{wang2019t}. Pre-training VAEs has been recently considered in conditional text generation to amortize the training of decoders and to allow easy adaptation in new generation tasks \cite{duan2019pre}.

Nevertheless, all aforementioned efforts utilize simple \textsc{RNN} \cite{hopfield1982neural} and shallow Transformer architectures thus they equip with limited model capacities. Pre-trained language models (PLMs) are recently introduced in VAEs to further boost both the generative and understanding abilities of VAEs. \citet{li2020optimus} proposed the first ``big VAE'' model at the same scale of \textsc{BERT} and GPT-2, their model connects two PLMs with different embedding spaces in a latent space, which demonstrate the efficacy for reducing related issue effectiveness in multiple NLP tasks. \citet{park2021finetuning} proposed to incorporate \textsc{T5} \cite{raffel2019exploring} into VAEs. \citet{fang2021transformer} utilized \textsc{GPT-2} \cite{radford2019language} in the VAE paradigm for controllable long story generation task. \citet{fang2022controlled} employed a discrete latent prior and additional noises to boost the control ability of a text VAE model. However, these methods essentially fine-tune two PLMs during model training, requiring a great amount of resources compared to \textsc{RNN}-based models or a single PLM.


\subsection{Parameter-Efficient PLMs}\label{sec:pe-methods}
For large pre-trained language models, massive training samples and huge parameter volumes help them gain unparalleled modeling ability. The conventional method to transfer a PLM to a specific data domain is fine-tuning, which mobilizes all parameters from the model. This can be intolerant in both computing and storage resources as the task load grows. Taming PLMs with high-efficiency w.r.t. distinct missions becomes one of the top trends in NLP. Various lightweight alternatives in PLMs were proposed. \citet{houlsby2019parameter} first came up with the idea to additionally add trainable adapter components with down sample and up sample layers in transformer blocks, which proved to achieve comparable results in NLP tasks with less than $10\%$ trainable parameters. Following this line, \citet{pfeiffer2020adapterfusion} improved the method by changing the adapter components to different positions of transformer blocks. \citet{li2021prefix} proposed to add trainable prefix to attention head in the model, while \citet{hu2021lora} created a shortcut in the attention domain of transformers which consists of trainable down and up sampling layers. Recently, \citet{he2021towards} concluded all these methods into a unified paradigm, which can be formalized as: $h\leftarrow \lambda_1 h+\lambda_2 \Delta a$,
where $h$ is the output of the attention layer or feedforward layer in one transformer block. The parameter $\lambda$ is varied according to different types of components, e.g., for Prefix tuning \cite{li2021prefix}, $\lambda_1 = 1 - \lambda_2$ with $\lambda_2$ to be a pre-assigned scalar. As for Adapter tuning \cite{houlsby2019parameter}, it does not employ a scalar, which means $\lambda_1 = \lambda_2 = 1$. Finally, the $\Delta a$ decides the receiving information for the current component from the previous layers. When $\Delta a$ is transformed from states ahead of the current attention block, the information is said to be ``parallelly'' shared, otherwise, it is ``sequantially'' shared with the model. These methods generally share a similar approach of introducing additional trainable parameters to PLMs instead of activating the original pre-trained transformers.

\section{\textsc{AdaVAE} Methodologies}
\begin{figure}[!t]
\centering
\includegraphics[width=0.75\linewidth]{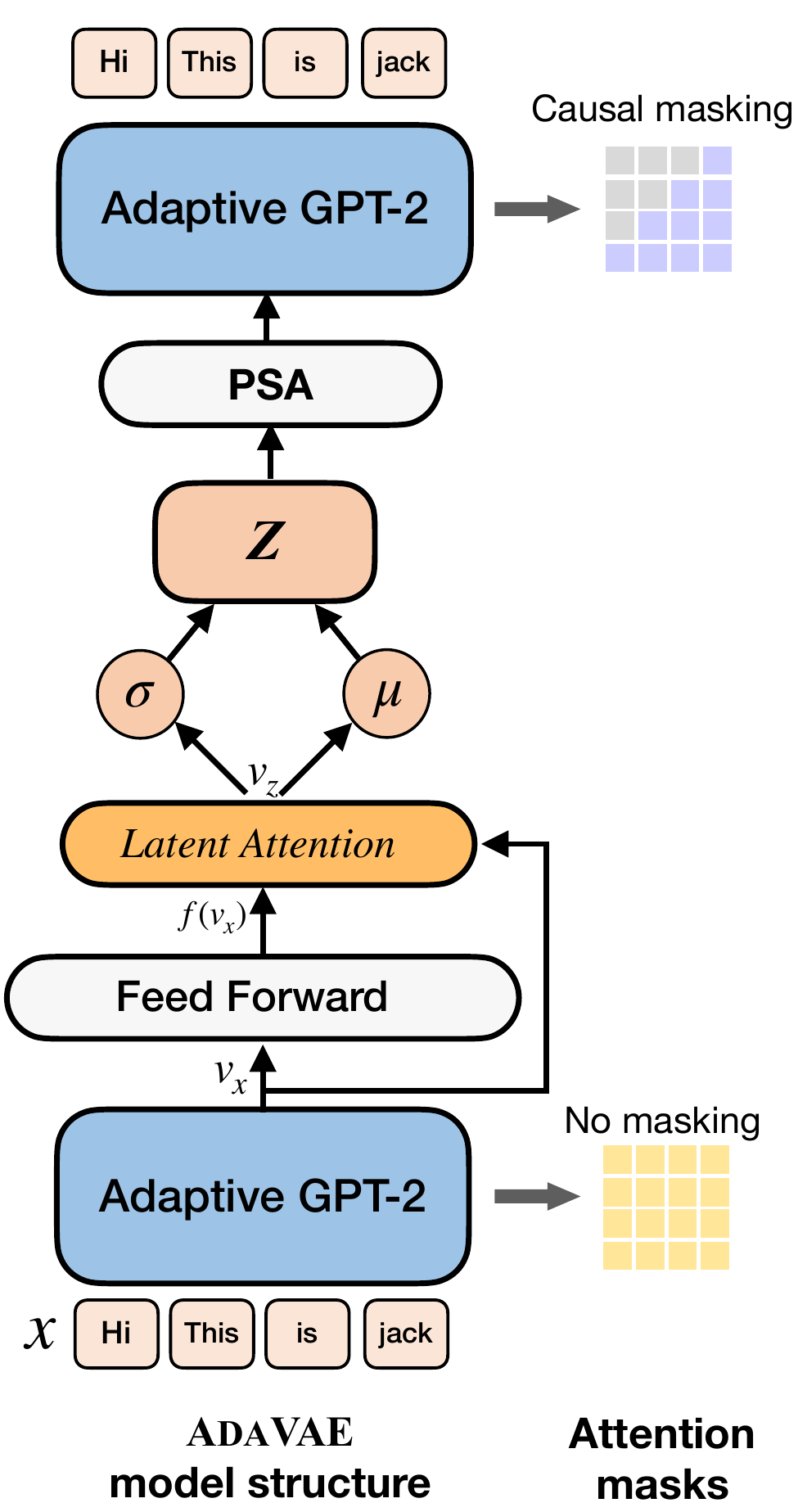}
\setlength{\abovecaptionskip}{0cm}
\caption{Model structure of \textsc{AdaVAE}. The model follows encoder-decoder architecture, where the encoder is an adaptive bi-directional \textsc{GPT-2} without attention mask and the decoder is an adaptive \textsc{GPT-2} with causal attention mask for autoregressive generation. $\mathcal{Z}$ is constructed by the proposed \textit{Latent Attention}. And the infusion method of $\mathcal{Z}$ is Pseudo Self Attention (\textbf{PSA}).}
\label{fig:modelstruct}
\end{figure}

In this section, we will detailly demonstrate our method from (1) encoder and decoder designing, (2) continuous latent space $\mathcal{Z}$ organization from encoder, and (3) latent knowledge infusion to decoder, are three core issues need to be considered when constructing a VAE-based model. Our overall model structure is shown in Figure~\ref{fig:modelstruct}.
\subsection{Adaptive GPT-2 Encoder and Decoder}
The encoder of a VAE should extract features from given contents to produce meaningful latent space, while the decoder ought to generate fluent sentences with given latent representations. In order to obtain a unified word embedding space in \textsc{AdaVAE}, we construct both encoder and decoder using GPT-2, which leaves us no worry about connecting two word embedding spaces from different models as in \cite{li2020optimus}. To make GPT-2 a qualified encoder, we take advice from mighty extractors such as \textsc{BERT}, one of its architectural advantages lies in the unmasked/bi-directional transformer layer. Thus we remove the causal mask in GPT-2 transformer layers to make it an encoder of \textsc{AdaVAE} with full vision of input contexts, this mask-free operation is widely used in the encoders of existing PLMs and VAEs \cite{raffel2019exploring,lewis2019bart,fang2021transformer}. As for the decoder, we employ GPT-2, a powerful generative transformer model by design.

The paradigm of fine-tuning two separate PLMs in large-scale VAEs requires a lot more computing resources than a single PLM, and the storage requirements will become too heavy to tolerate as the task loads increase. To avoid such dilemma, we propose and explore different parameter-efficient components including different types or insertion methods into encoder and decoder layers, which means only additional minimum parameters need to be activated for every task. Specifically, we propose a parallel adapter placed after the feedforward layers (see Section \ref{sec:pe-methods}) of each attention block in both encoder\&decoder for \textsc{AdaVAE}. And we further compare our approach with different adapters and Prefix tuning method \cite{li2021prefix} for ablation study in Section \ref{sec:lm}. Overall, these two settings make \textsc{AdaVAE} \textbf{more elegant} to be constructed and \textbf{more efficient} to be trained compared with existing ``big VAEs''.

\subsection{From Encoder to Latent Space $\mathcal{Z}$}
How to form the latent space from the encoder and utilize it in the decoder to narrow the gap between discrete sentences to the continuous latent embedding is a key problem. \citet{li2020optimus} and \citet{park2021finetuning} employed a pooled feature from the last encoder layer and pass it to a linear transformation to obtain latent space, which may be not sufficient to leverage the knowledge learned from transformer layer. \citet{fang2021transformer} used the last state from the encoder as both the key and value vectors to conduct averaged attention by matrix multiplication. Their model learns both prior and posterior of the latent space from the same type of input (i.e., the same textual content), and shares most of the learning parameters in this process. We contend that 1. vectors from different domains in the attention operation (i.e., key, value, query) should be distinct to carry specific knowledge. 2. To avoid potential KL collapse issue, one ought to produce latent posterior and prior from different types of input sources \cite{lucas2019understanding}. 

We thus propose the improved \textit{Latent Attention} operation to generate meaningful latent space in \textsc{AdaVAE}: to get latent vector $\boldsymbol{v_z}$, we adopt the last hidden state $\boldsymbol{v_x}$ from the encoder and assign:
\begin{equation}
    \begin{aligned}
    &\mathbf{Q_z} = \mathbf{E}, \mathbf{K_z} = f(\boldsymbol{v_x}), \mathbf{V_z} = \boldsymbol{v_x},\\
    &\boldsymbol{v_z} = Attention(\mathbf{Q_z}, \mathbf{K_z}, \mathbf{V_z}),
    \end{aligned}
\end{equation}
where $\mathbf{E}$ is a identity matrix with the same size of $\boldsymbol{v_x}$, $f(\cdot)$ is a linear transformation for mapping $\boldsymbol{v_x}$ to the key vector space, and finally the $\boldsymbol{v_z}$ is from the attention operation between derived $\mathbf{Q_z, K_z, V_z}$. Then the latent vector $\boldsymbol{v_z}$ is taken to reparameterize the mean ($\boldsymbol{\mu}$) and variance ($\boldsymbol{\sigma}$) of $\mathcal{Z}$:
\begin{equation}
    \begin{aligned}
    & \boldsymbol{\mu} = f_{\mu}(\boldsymbol{v_z}), \log (\boldsymbol{\sigma}) = f_{\sigma}(\boldsymbol{v_z}),\\
    & \boldsymbol{z} = \boldsymbol{\mu} + \boldsymbol{\sigma} \odot \epsilon,\quad \epsilon\sim \mathcal{N}(0,I),
    \end{aligned}
\end{equation}
where $\boldsymbol{z}$ is a latent vector sampled from space $\mathcal{Z}$, $f_{\mu}$ and $f_{\sigma}$ are two linear transformations, $\odot$ is the element-wise multiplication. Note that, we only use it to model the posterior of latent space and leave its prior to be a normalized Gaussian \cite{bowman2015generating}. This setting takes full advantage of learned information from the encoder and reduces the possibility of KL vanishing problem that may occur in the previous work \cite{fang2021transformer}.

\begin{figure}[!t]
\centering
\includegraphics[width=1\linewidth]{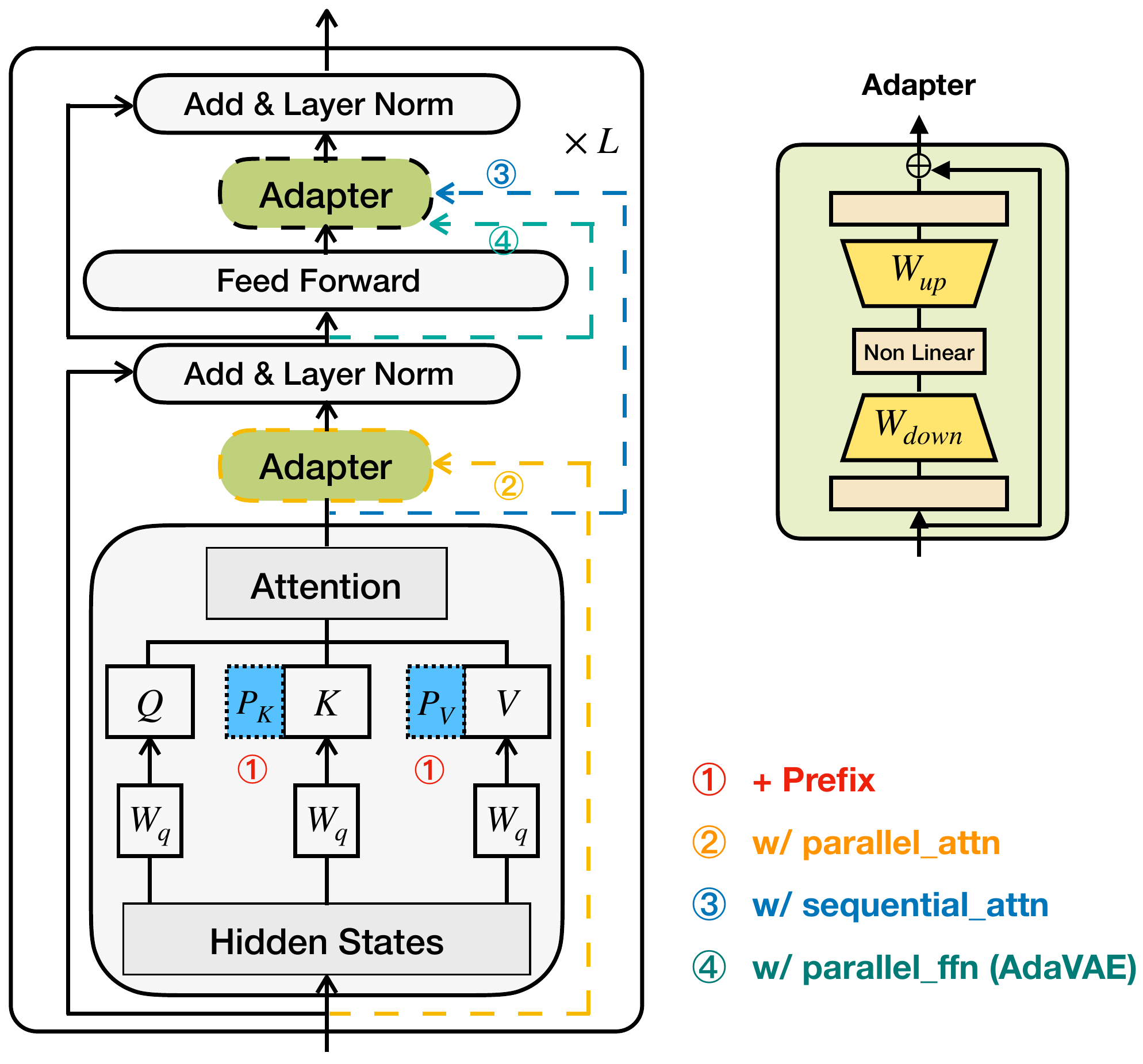}
\setlength{\abovecaptionskip}{0cm}
\caption{Different parameter-efficient components in the proposed model. Dashed lines with different colors represent the shorcut connection from different positions in one transformer block. We introduce these components, and explore their respect performance in Section \ref{sec:lm}.}
\label{fig:pe-components}
\end{figure}

\subsection{From Latent Space $\mathcal{Z}$ to Decoder}
The way to infuse learned latent knowledge from the latent space into the generative process determines the effectiveness of the decoder employing learned representations. 
Inspired by existing methods, we investigate two different frames to add latent variables into decoder layers. For a latent variable $\boldsymbol{z}\in \mathbb{R}^d$ drawn from $\mathcal{Z}$, we investigate two different fusion methods:
\begin{itemize}
    \item \textbf{Add to Memory (AtM)} \cite{li2020optimus} projects $\boldsymbol{z}$ to both attention key and value spaces by a unified linear layer $f(\cdot)$, and concatenate them with key and value vector in each attention layer:
\begin{equation}
\centering
    \begin{aligned}
    &\boldsymbol{k_z} = \boldsymbol{v_z} = f(\boldsymbol{z})\in \mathbb{R}^{d\times l},\\&
    \boldsymbol{k = [k; k_z]}\in \mathbb{R}^{l}, \boldsymbol{v = [v; v_z]}\in \mathbb{R}^{l},
    \end{aligned}
\end{equation}
where $l, \boldsymbol{k}, \boldsymbol{v}$ are the size of both key and value spaces, Key vector and value vector severally.
    \item \textbf{Pseudo Self-Attention (PSA)} \cite{fang2021transformer} shares a similar idea with AtM, but it uses separate convolutional transformations with $\boldsymbol{z}$ as input to make sure that $k_z \not= v_z$, then PSA concatenates them with respect vectors just like AtM to conduct the pseudo self-attention in decoder layers.
\end{itemize}

\subsection{Model Training}
The training loss of our model is based on the plain VAE. To maximize the potential of our model, we further incorporate free bit threshold and KL annealing techniques during model training.
\subsubsection{Free Bit Threshold}
The core idea of free bit (FB) thresholding is to mitigate the useless influence of meaningless latent dimensions by replacing the KL term in \textbf{ELBO} (as in Eq. (\ref{eq:elbo})) with a hinge loss term that maxes each latent unit. We followed previous VAE-related works \cite{li2019surprisingly,pelsmaeker2019effective} to apply the free bit threshold to the entire KL term:
\begin{equation}
  \mathcal{L}_{\text{KL}} = \max \left[\lambda, \sum_{i} D_{\mathrm{KL}}\left(q_{\boldsymbol{\phi}}\left(z_{i} \mid \mathbf{x}\right) \| p\left(z_{i}\right)\right)\right],
\end{equation}
where $z_i$ is the $i$th dimension in the latent representation. Note that, different from \textsc{Optimus}, which sets thresholds to each dimension of the latent space, we set an overall threshold on $\mathcal{Z}$ to stabilize the training procedure of our model. Finally, the overall training loss of the proposed model is 
\begin{equation}
    \begin{aligned}
    & \mathbb{E}_{q(\boldsymbol{z} \mid \boldsymbol{X})}\left[\log p(\boldsymbol{X} \mid \boldsymbol{z}\right)]\\&-\beta \max \left[\lambda, \sum_{i} D_{\mathrm{KL}}\left(q_{\boldsymbol{\phi}}\left(z_{i} \mid \mathbf{x}\right) \| p\left(z_{i}\right)\right)\right],
    \end{aligned}
\end{equation}
where $\beta, \lambda$ are two hyper-parameters to be tuned in training.
\subsubsection{KL Annealing}
One simple way to alleviate the KL collapse problem is annealing the KL weight during model training. Generally speaking, the weight $\beta$ of KL divergence in the VAE objective gradually changes from 0 to 1. This trick can be applied once \cite{bowman2015generating} or multiple times \cite{fu2019cyclical} during training. We employed cyclic annealing with 4 cycles for KL weight $\beta$.
\section{Experimental Details}
In this section, we will introduce datasets, techniques that we employed in our model as well as specific model architecture details.

\begin{table}[!t]
\centering
\setlength\tabcolsep{4pt}
\normalsize
\begin{tabular}{c|cccc} 
\toprule[1.5pt]
Dataset         & \# Train & \# Val & \# Test & Avg. Len.  \\ 
\midrule
\texttt{YELP}            & 100K     & 10K    & 10K     & 96              \\
\texttt{YAHOO}           & 100K     & 10K    & 10K     & 79              \\
\texttt{PTB}             & 42K      & 3.0K     & 3.0K      & 21              \\
$\texttt{YELP}_S$ & 44K      & 10K    & 10K     & 9.0             \\
\texttt{WNLI}            & 0.64K      & 0.07K     & 0.15K     & 27              \\
\texttt{SST-2}           & 67K      & 0.9K   & 2.0K      & 9.4             \\
\bottomrule[1.5pt]
\end{tabular}
\caption{Statistics of datasets. We present the size of train/val/test sets and the average length for 6 datasets.}
\label{tab:dataset}
\end{table}

\subsection{Implementation Details}
\begin{table*}
\setlength\tabcolsep{7pt}
\renewcommand\arraystretch{1}
\centering
\small
\begin{tabular}{l|cccc|cccc|c} 
\toprule[1.5pt]
Dataset       & \multicolumn{4}{c|}{\texttt{YELP }}                                                                   & \multicolumn{4}{c|}{\texttt{YAHOO}}                                                                  & \multirow{3}{*}{\textbf{\#params.}}  \\\cline{1-9}
                                                           & \multicolumn{2}{c}{LM}                      & \multicolumn{2}{c|}{Repr.}          & \multicolumn{2}{c}{LM}                      & \multicolumn{2}{c|}{Repr.}          &                                    \\
Method                                            & \textbf{PPL}$\downarrow$ & \textbf{-ELBO}$\downarrow$ & \textbf{MI}$\uparrow$ & \textbf{AU}$\uparrow$ & \textbf{PPL}$\downarrow$ & \textbf{-ELBO}$\downarrow$ & \textbf{MI}$\uparrow$ & \textbf{AU}$\uparrow$ &                                    \\ 
\midrule
\textsc{AdaVAE}                                            & 15.49                    & 125.56                     & 7.55                  & 32                    & 14.23                    & 121.40                     & 7.49                  & 32                    & $14.66\%$                          \\ 
\textbf{w/ Fine-tuning}                                    & 18.59                    & 125.02                     & 7.49                  & 32                    & 15.57                    & 121.05                     & 7.52                  & 32                    & $100.00\%$                            \\
\midrule
\textbf{w/ LG}                                             & 31.01                    & 129.17                     & 3.32                  & 32                    & 19.64                    & 123.16                     & 2.32                  & 32                    & $15.13\%$                          \\
\textbf{w/ LG; +AtM}                              & 30.44                    & 129.39                     & 4.29                  & 32                    & 22.54                    & 122.19                     & 4.36                  & 32                    & $20.82\%$                          \\ 
\midrule
\textbf{+Prefix}                                         & 14.96                    & 124.13                     & 6.55                  & 32                    & 15.17                    & 120.89                     & 3.70                  & 32                    & $15.65\%$                          \\ 
\textbf{w/ parallel\_attn}                                 & 16.32                    & 125.91                     & 7.57                  & 32                    & 15.22                    & 122.22                     & 7.40                  & 32                    & $14.66\%$                          \\
\textbf{w/ sequential\_attn}                               & 17.98                    & 127.33                     & 7.55                  & 32                    & 15.05                    & 121.69                     & 7.47                  & 32                    & $14.66\%$                          \\
\bottomrule[1.5pt]
\end{tabular}
\caption{The proposed VAE-based model \textsc{AdaVAE} with different parameter-efficient/latent generation frameworks on language modeling task. \textbf{\#params.} is the percentage of (additional) training parameters compared with the original language model. The $\lambda = 0.50$ in all cases.}
\label{tab:ablation}
\end{table*}

For model architecture and initialization, the encoder and decoder were separately 8 layers and 12 layers GPT-2 transformers initialized from pre-trained weights in Hugginface.\footnote{\url{https://huggingface.co/gpt2}} The parameter efficient components were chosen from Adapter \cite{houlsby2019parameter} and Prefix \cite{li2021prefix}, the hidden size of the Adapter was chosen to be 128 or 512 depending on different tasks, while the hidden size of the Prefix was 30 for the ablation study in language modeling task. The hidden size of latent space $\mathcal{Z}$ was set to 32 for language modeling and 768 for classification and generation.

For training details, we first activated the parameter-efficient components in the encoder and parameters in latent spaces for the first $1/6$ training steps and then added parameter-efficient components in the decoder for the rest of the training time, this setting is helpful for training a VAE model \cite{li2019surprisingly}. Similarly, we also employed linear warming-up procedure for learning rate \cite{popel2018training} to make it increases linearly from 0 to $5\times 10^{-5}$ in the first $1/6$ training iterations. We train the model with the batch size of around 64 on one TITAN X GPU with 12G memory.

\subsection{Datasets Details}
The detailed dataset statistics are in TABLE \ref{tab:dataset}. We conduct three generation tasks and an understanding task span from six datasets. For language modeling task, we use \texttt{YELP, YAHOO} and \texttt{PTB} from \textsc{Optimus} \cite{li2020optimus} directly. For controllable generation, we take a shorter version of \texttt{YELP} dataset (denoted as $\texttt{YELP}_S$), which is originally designed for the style transfer learning with labels \cite{shen2020educating}. As for style transfer task, we use the same $\texttt{YELP}_S$ dataset as in the mentioned generation task. For language classification (understanding) task, we apply $\texttt{YELP}_S$ data as well as additionally introduced \texttt{WNLI} and \texttt{SST-2} dataset from GLUE benchmark \cite{wang2018glue}.

\section{Experimental Results and Analysis}
In this section, we focus on validating experiments that explain the generative ability and feature extraction ability of the proposed model. In addition, we will detailly introduce respect baselines and evaluation metrics in each task.
\subsection{Language Modeling Ability} \label{sec:lm}
For the evaluation of language modeling ability,  we took \textit{Perplexity} (\textbf{PPL}), the negative evidence lower bound (\textbf{ELBO}), mutual information between input sentence and $\mathcal{Z}$ (\textbf{MI}) and activated units in the latent space (\textbf{AU}) as measurements. All metrics were implemented exactly following the public codebase\footnote{\url{https://github.com/ChunyuanLI/Optimus}} for fair comparisons. While \textbf{PPL} and \textbf{ELBO} measure the fluency of generated sentences, \textbf{MI} and \textbf{AU} indicate the representation learning capacity of a latent variable model.

We first explore the effects of different types of parameter-efficient components as well as latent space generation\&infusion methods for the proposed model. We explored 7 types of VAE models in Table~\ref{tab:ablation}, their visual illustrations are in Figure~\ref{fig:pe-components}: 
\begin{enumerate}
    \item \textsc{AdaVAE}: Uses the proposed parallel adapter for feedforward layer in transformers, \textit{Latent Attention} for latent space construction, \textbf{PSA} for representation infusion in the decoder. 
    \item \textbf{w/ Fine-tuning}: Removes adapters and fine-tunes the original model.
    \item \textbf{w/ LG}: Uses the pooled feature from encoder and a linear layer to form $\mathcal{Z}$.
    \item \textbf{w/ LG; +AtM}: Uses LG and both the \textbf{PSA\&AtM} methods together for latent infusion.
    \item \textbf{+Prefix}: Adds Prefix components to attention blocks.
    \item \textbf{w/ parallel\_attn}: Replaces the original adapters with parallel adapters for attention outputs.
    \item \textbf{w/ sequential\_attn}: Replaces the original adapters with sequential adapters for attention outputs.
\end{enumerate}

Table~\ref{tab:ablation} shows the proposed model with different adding components or training schemes, \textsc{AdaVAE} achieves a good trade-off between language modeling and representation learning ability among presented models. Focusing on specific model structures, (1) From the first row in Table~\ref{tab:ablation}, fine-tuning the model does not bring significant improvement with significantly larger training parameters ($100\%$ vs. $14.66\%$), and even perform worse on \textbf{PPL} (\texttt{YELP} and \texttt{YAHOO}) and one \textbf{MI} metric (\texttt{YELP}) compared with the proposed parameter-efficient training method. This demonstrates our design of adaptive \textsc{GPT-2} encoder\&decoder in \textsc{AdaVAE} is efficient, which leads the way to resolve the excessive resource consumption problem faced by existing ``big VAEs''. (2) At the second row, we could find that replacing the proposed \textit{Latent Attention} with \textbf{LG} from \textsc{Optimus} makes the overall model performance worse. This demonstrates the unfitness to employ simple linear transformation on learned transformer features for space $\mathcal{Z}$. Adding \textbf{AtM} to infuse latent representation with \textbf{PSA} generally boosts model's representation learning with higher \textbf{MI} scores. This is ascribed to the cumulative use of learned latent knowledge with added trainable parameters ($5\%$ more model parameters) with \textbf{AtM} and \textbf{PSA}. (3) At the last row in Table~\ref{tab:ablation}, we focus on the gain of model performance brought by different parameter-efficient components. Adding the Prefix component will introduce more training parameters, but with little or even negative performance gain in the learning process (lower \textbf{MI} and higher \textbf{PPL} on \texttt{YAHOO}). Besides, replacing the original parallel adapters for the feedforward layers with either parallel or sequential adapters for attention output lags behind the proposed adapter in the model. These help us to understand the adaptive tuning method in our proposed model is practical to use. Overall, \textsc{AdaVAE} with proposed adaptive \textsc{GPT-2}s and the \textit{Latent Attention} shows more consistent performance in the capacity trade-off among ablation components.

We further compare our model with different baselines from conventional VAEs with state-of-the-art PLM-based VAEs. Our baseline models are listed as follows:
\begin{itemize}
    \item \textsc{iVAE} \cite{fang2019implicit}: a VAE model considers implicit posterior representation instead of the explicit form.
    \item \textsc{GPT-2} \cite{radford2019language}: a large-scale LM pre-trained on large scale real-world dataset and fine-tuned on each dataset for 1 epoch.
    \item \textsc{T5 VAE} \cite{park2021finetuning}: a PLM-based VAE model that fine-tunes the encoder and decoder of T5 \cite{raffel2019exploring} model into VAE setting.
    \item \textsc{Optimus} \cite{li2020optimus}: the first PLM-based VAE model that connects pre-trained \textsc{\textsc{BERT}} and \textsc{GPT-2} by organizing a continuous latent space.
    \item \textsc{DPrior} \cite{fang2022controlled}: a PLM-based VAE model uses discrete latent prior and additional noise in the latent space to strengthen control ability under \textsc{Optimus}.
\end{itemize}

\begin{table*}[!t]
\setlength\tabcolsep{2pt}
\renewcommand\arraystretch{1}
\small
\centering
\begin{tabular}{cl|cccc|cccc|cccc} 
\toprule[1.5pt]
Dataset                               &                     & \multicolumn{4}{c|}{\texttt{PTB }}                                      & \multicolumn{4}{c|}{\texttt{YELP }}                                     & \multicolumn{4}{c}{\texttt{YAHOO }}                                      \\ 
\midrule
                                               &                     & \multicolumn{2}{c}{LM} & \multicolumn{2}{c|}{Repr.} & \multicolumn{2}{c}{LM} & \multicolumn{2}{c|}{Repr.} & \multicolumn{2}{c}{LM} & \multicolumn{2}{c}{Prepr.}  \\
Method                                &                     & \textbf{PPL}$\downarrow$ & \textbf{-ELBO}$\downarrow$    & \textbf{MI}$\uparrow$ & \textbf{AU}$\uparrow$            & \textbf{PPL}$\downarrow$ & \textbf{-ELBO}$\downarrow$    & \textbf{MI}$\uparrow$ & \textbf{AU}$\uparrow$            & \textbf{PPL}$\downarrow$ & \textbf{-ELBO}$\downarrow$    & \textbf{MI}$\uparrow$ & \textbf{AU}$\uparrow$             \\ 
\midrule
                                              
& \textsc{iVAE}     & 53.44        & 87.20            &          & 32                     & 36.88        &        348.70           &          & 32                     & 47.93        & 309.10            &          & 32                      \\ 
                                               & \textsc{GPT-2}      & 24.23        & -                 & -           & -                      & 23.40        & -                 & -           & -                      & 22.00        & -                 & -           & -                       \\
                                               & \textsc{LSTM-LM}    & 100.47       & 101.04            & -           & -                      & 42.60        & 358.10            & -           & -                      & 60.75        & 328.00            & -           & -                       \\ 
& \textsc{T5 VAE}  & 57.69 & 101.17 & & 11 & 53.05 & 166.15 & 5.55 & 10 & 54.40 & 140.57 & 5.43 & 28 \\ 
& \textsc{DPrior} & 14.74 & \textbf{72.84} &   & 32 & \textbf{14.52} & 287.92 &  & 32 & 14.67 & 244.01 &   & 32 \\ \midrule
\multirow{5}{*}{\textsc{Optimus}}   & $\lambda=0.05$      & 23.58        & 91.31             & 3.78        & 32                     & 21.99        & 337.41            & 2.54        & 32                     & 22.34        & 282.70            & 5.34        & 32                      \\
                                               & $\lambda=0.10$      & 23.66        & 91.60             & 4.29        & 32                     & 21.99        & 337.61            & 2.87        & 32                     & 22.56        & 289.88            & 5.80        & 32                      \\
                                               & $\lambda=0.25$      & 24.34        & 93.18             & 5.98        & 32                     & 22.20        & 340.03            & 5.31        & 32                     & 22.63        & 290.69            & 7.42        & 32                      \\
                                               
                                               & \cellcolor[rgb]{0.929,0.929,0.929}{$\lambda=0.50$}      & \cellcolor[rgb]{0.929,0.929,0.929}{26.69}        & \cellcolor[rgb]{0.929,0.929,0.929}{96.82}             & \cellcolor[rgb]{0.929,0.929,0.929}{7.64}        & \cellcolor[rgb]{0.929,0.929,0.929}{32}                     & \cellcolor[rgb]{0.929,0.929,0.929}{22.79}        & \cellcolor[rgb]{0.929,0.929,0.929}{344.10}            & \cellcolor[rgb]{0.929,0.929,0.929}{7.67}        & \cellcolor[rgb]{0.929,0.929,0.929}{32}                     & \cellcolor[rgb]{0.929,0.929,0.929}{23.11}        & \cellcolor[rgb]{0.929,0.929,0.929}{293.34}            & \cellcolor[rgb]{0.929,0.929,0.929}{8.85}        & \cellcolor[rgb]{0.929,0.929,0.929}{32}                      \\
                                               & $\lambda=1.00$      & 35.53        & 77.65             & \textbf{8.18}        & 32                     & 24.59        & 353.67            & \textbf{9.13}        & 32                     & 24.92        & 301.21            & \textbf{9.18}        & 32                      \\ 
\midrule
\multirow{5}{*}{\textsc{AdaVAE}}    & $\lambda=0.05$      & 23.18        & 89.27             & 1.21        & 32                     & 31.22        & \textbf{\textcolor{blue}{115.74}}            & 1.07        & 32                     & 26.53        & \textbf{\textcolor{blue}{109.69}}            & 1.20        & 32                      \\
                                               & $\lambda=0.10$      & 18.94        & \textcolor{blue}{88.50}             & 2.14        & 32                     & 27.87        & 116.66            & 2.21        & 32                     & 23.69        & 110.21            & 2.17        & 32                      \\
                                               & $\lambda=0.25$      & \textbf{\textcolor{blue}{11.97}}        & 89.52             & 5.54        & 32                     & 18.21        & 116.62            & 6.02        & 32                     & 16.04        & 112.39            & 5.88        & 32                      \\
                                               
                                               &  \cellcolor[rgb]{0.929,0.929,0.929}{$\lambda=0.50$}      & \cellcolor[rgb]{0.929,0.929,0.929}{12.77}        & \cellcolor[rgb]{0.929,0.929,0.929}{99.46}             & \cellcolor[rgb]{0.929,0.929,0.929}{7.54}        & \cellcolor[rgb]{0.929,0.929,0.929}{32}                     & \cellcolor[rgb]{0.929,0.929,0.929}{\textcolor{blue}{15.49}}        & \cellcolor[rgb]{0.929,0.929,0.929}{125.56}            & \cellcolor[rgb]{0.929,0.929,0.929}{7.55}        & \cellcolor[rgb]{0.929,0.929,0.929}{32}                     & \cellcolor[rgb]{0.929,0.929,0.929}{\textbf{\textcolor{blue}{14.23}}}        & \cellcolor[rgb]{0.929,0.929,0.929}{121.40}            & \cellcolor[rgb]{0.929,0.929,0.929}{7.49}        & \cellcolor[rgb]{0.929,0.929,0.929}{32}                      \\
                                               & $\lambda=0.75$      & 27.98        & 110.35            & \textcolor{blue}{7.82}        & 32                     & 35.92        & 139.46            & \textcolor{blue}{7.62}        & 32                     & 31.01        & 136.06            & \textcolor{blue}{7.65}        & 32                      \\
\bottomrule[1.5pt]
\end{tabular}
\caption{Language modeling ability of different VAE-based models. Best values of the proposed model and all models are in \textcolor{blue}{blue} and \textbf{boldface} respectively. $\lambda = 0.5$ is a good choice for \textsc{AdaVAE} that perform better in \textbf{LM} ability and slightly worse in \textbf{MI} measurement compared with $\textsc{Optimus}$.}
\label{tab:lm}
\end{table*}

Table~\ref{tab:lm} shows the comparison between the proposed model and some SOTA VAEs. We draw the following conclusions based on it: firstly, to find the best value for every metric on each dataset, the proposed model holds the lowest \textbf{PPL} and \textbf{-ELBO} values on two datasets among all baselines, demonstrating the advantages in language modeling of our proposed approach. Though the performance of proposed model surges ahead of most baselines on all metrics, there still remain a small margin between our model and \textsc{Optimus} on \textbf{MI} scores (less than 0.15 when $\lambda = 0.5$ on \texttt{PTB}, \texttt{YELP}). We argue that it is essential to consider both LM and representation-related statistical results simultaneously. \textsc{Optimus} achieves the highest \textbf{MI} score with much worse \textbf{PPL} or \textbf{-ELBO} results compared with \textsc{AdaVAE}, demonstrating that \textsc{Optimus} actually sacrifices large amount of LM ability to gain some improvements on \textbf{MI}. While our model stays a steady trade-off in this circumstance (generally better \textbf{PPL} and \textbf{-ELBO} and slightly lower \textbf{MI}). Also note that only \textsc{AdaVAE} \textit{activates partial model parameters during training}, which is another huge advantage compared to strong fine-tuned baselines (e.g., \textsc{DPrior}, \textsc{Optimus}).

Then we focus on the effect of $\lambda$. As the free bits threshold $\lambda$ increases, \textbf{MI} value generally yields a better performance in both \textsc{Optimus} and \textsc{AdaVAE}. This is because a larger KL threshold brings up the amount of knowledge that should be learned in the latent space. While the trend of \textbf{PPL} value is monotonous with $\lambda$ in \textsc{Optimus}, there is a rebound in \textsc{AdaVAE} with the optimal \textbf{PPL} with $\lambda = 0.50$ on two corpus. With a fairly high \textbf{MI} score, we believe it is the suitable $\lambda$ value for a good trade-off between model's language modeling and representation learning ability.

\subsection{Controllable Text Generation}
We further conduct controllable generation on $\texttt{YELP}_S$ dataset collected by \citet{shen2020educating}, which contains 444K training sentences, and we use separated datasets of 10k sentences for validation/testing respectively as in \textsc{Optimus}. The goal is to generate text reviews given the positive/negative sentiment from the $\texttt{YELP}_S$ dataset. For evaluation metrics, we employed the following metrics for automatic evaluation: Accuracy (\textbf{Acc.}) is measured by a pre-trained classifier on the $\texttt{YELP}_S$ dataset, indicating the controllability of the model. BLEU (\textbf{B.}) for the quality evaluation of generated sentences. G-score (\textbf{G-S.}) computes the geometric mean of Accuracy and BLEU, measuring the comprehensive quality of both content and style. Self-BLEU (\textbf{S-B.}) measures the diversity of the generated sentences. And for BLEU-F1 (\textbf{B-F1}), we have: $\text{BLEU-F1} = \frac{2\times \text{BLEU}\times (1 - \text{Self-BLEU})}{\text{BLEU} + (1 - \text{Self-BLEU})}$,
which evaluates the overall metric involving text quality and diversity simultaneously. 

\begin{table}[]
\small
\centering
\setlength\tabcolsep{2pt}
\begin{tabular}{l|ccccc}
\toprule[1.5pt]
Model       & \textbf{Acc. $\uparrow$} & \textbf{B. $\uparrow$} & \textbf{G-S. $\uparrow$} & \textbf{S-B. $\downarrow$} & \textbf{B-F1$\uparrow$} \\ \midrule
\textsc{Control-Gen}  & 0.878                       & 0.389                   & 0.584                      & 0.412                          & 0.468                      \\
\textsc{ARAE}        & 0.967                       & 0.201                   & 0.442                      & 0.258                          & 0.316                      \\
\textsc{NN-Outlines} & 0.553                       & 0.198                   & 0.331                      & 0.347                          & 0.304                      \\
\textsc{Optimus}     & \textcolor{blue}{\textbf{0.998}}                       & 0.398                   & 0.630                      & \textcolor{blue}{\textbf{0.243}}                 & 0.522                      \\\midrule
\textsc{AdaVAE}      & 0.889                       & 0.317                   & 0.531                      & 0.565                          & 0.367                      \\
\cellcolor[rgb]{0.929,0.929,0.929}{$\textsc{AdaVAE}_{dec}$}      & \cellcolor[rgb]{0.929,0.929,0.929}{0.931}                       & \cellcolor[rgb]{0.929,0.929,0.929}{\textcolor{blue}{\textbf{0.608}}}                   & \cellcolor[rgb]{0.929,0.929,0.929}{\textcolor{blue}{\textbf{0.752}}}                      & \cellcolor[rgb]{0.929,0.929,0.929}{0.498}                          & \cellcolor[rgb]{0.929,0.929,0.929}{\textcolor{blue}{\textbf{0.550}}}                      \\
\bottomrule[1.5pt]
\end{tabular}
\caption{Controllable text generation on $\text{YELP}_S$, the best statistics are in \textcolor{blue}{\textbf{blue}}.}
\label{tab:controlgen}
\end{table}

\begin{table}[!t]
\small
\setlength\tabcolsep{2pt}
\begin{tabular}{l|l}
\toprule[1.5pt]
\textbf{Negative}                            & \textbf{Positive}                     \\ \midrule
i'm not going back.                 & i've had a great experience. \\
\begin{tabular}[c]{@{}l@{}} this salad's terrible.\end{tabular}              & i'm definitely coming back.  \\
\begin{tabular}[c]{@{}l@{}} i'm not sure if the fries\\taste.\end{tabular}     & i've had the perfect veggie! \\
\begin{tabular}[c]{@{}l@{}} i've never been to this\\restaurant.\end{tabular}  & i'm always happy.            \\
\begin{tabular}[c]{@{}l@{}} i'm not impressed with\\the food!\end{tabular}    & i've had the best pizza.     \\
\bottomrule[1.5pt]
\end{tabular}
\caption{Generated texts on $\text{YELP}_S$ with sentiment labels.}
\label{tab:controlgen_sent}
\end{table}

The baselines are described as follows:
\begin{itemize}
    \item \textsc{Control-Gen} \cite{hu2017toward}: employs discriminator on latent space to control the generation process.
    \item \textsc{ARAE} \cite{zhao2018adversarially}: learns an auto-encoder first, and then train a \textsc{GAN} to produce the latent vectors. 
    \item \textsc{NN-Outlines} \cite{subramanian2018towards}: uses a general purpose encoder for text generation, and a second stage training with labeled examples for the controllable generation task.
    \item \textsc{Optimus} \cite{li2020optimus}: finetunes the model for text general generation, then conducts second training stage with a conditional \textsc{GAN} \cite{mirza2014conditional} on latent space based on given labels.
\end{itemize}

To verify our model's capacity on the task of controllable generation, we took a similar experimental setup as \textsc{ARAE} and \textsc{Optimus}: we first trained our \textsc{AdaVAE} on $\texttt{YELP}_S$ dataset without label information, then we employed a conditional \textsc{GAN} on the learned latent representation of our model to produce latent vector $z_y$ based on label $y$. And finally, we explored two types of training settings for the decoder to produce controllable texts, i.e., \textsc{AdaVAE} only activates parameter-efficient adapters in the decoder during generation, which equips $12.03\%$ trainable parameters. $\textsc{AdaVAE}_{dec}$ activates all parameters in the decoder during generation, which equips $46.79\%$ trainable parameters compared with the original LM.

\begin{table}[!t]
\centering
\setlength\tabcolsep{1pt}
\centering
\small
\begin{tabular}{l|c|ccc|c} 
\toprule[1.5pt]
\multicolumn{2}{c|}{Model}                                                                                   & \texttt{WNLI} & $\texttt{YELP}_S$ & \texttt{SST-2} & \multirow{2}{*}{\textbf{\#param.}}  \\
\multicolumn{2}{c|}{\text{Dataset Size }}                                                                             & 0.63K  & 44K  & 67K   &                                      \\ 
\hline\hline
\multirow{3}{*}{\textbf{FB}}    & \textsc{BERT}                                                             & 0.577         & $\leq$0.88    & 0.731          & -                                    \\
                                            & \textsc{Optimus}                                                    & 0.563         & $\leq$0.92    & 0.789          & -                                    \\
\midrule
\multirow{3}{*}{\textbf{FT}}      & \textsc{BERT}                                                             & 0.544         & 0.984         & 0.923          & $100.0\%$                              \\
                                            & \textsc{Optimus}                                                    & 0.563         & $\leq$0.98    & 0.924          & $100.0\%$                              \\
                                            & \textsc{AdaVAE}                                                           & 0.586         & 0.968         & 0.860          & $100.0\%$                              \\ 
\midrule
\multirow{5}{*}{\textbf{PE}} & \textsc{BERT} ($\alpha=128$)                                             & 0.524         & 0.965         & 0.902          & $4.41\%$                             \\
                                            & \textsc{BERT} ($\alpha=512$)                                              & 0.531         & 0.973         & 0.911          & $17.49\%$                            \\
                                            & \textsc{AdaVAE} ($\alpha=128$)                                            & 0.563         & 0.966         & 0.853          & $2.90\%$                             \\
                                            & \textsc{AdaVAE} ($\alpha=512$)                                            & 0.589         & 0.961         & 0.840          & $7.82\%$                             \\
\bottomrule[1.5pt]
\end{tabular}
\caption{Latent classification accuracy on datasets with varied sizes of training examples. \textbf{FB, FT} and \textbf{PE} represent feature-based tuning, fine-tuning and parameter-efficient training with adapters respectively.}
\label{tab:cls}
\end{table}

\begin{figure}[!t]
\centering
\includegraphics[width=1\linewidth]{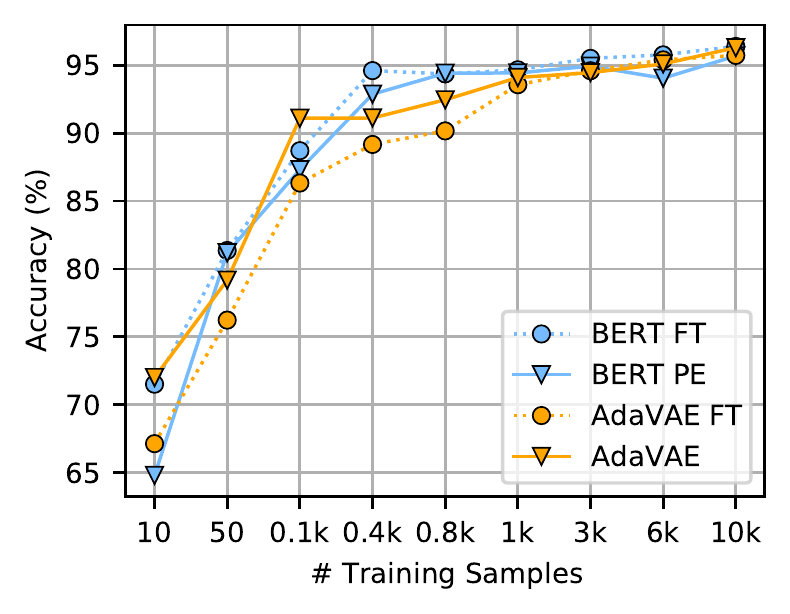}
\setlength{\abovecaptionskip}{0cm}
\caption{Testing accuracy with a varying number of total labeled training samples on $\text{YELP}_S$ dataset.}
\label{fig:lowcls}
\end{figure}

\newcommand{\tsnesize}{0.31\linewidth}
\begin{figure*}[ht]
\centering
\subfigure[\textsc{AdaVAE} with $\alpha=128$]{
\includegraphics[width=\tsnesize]{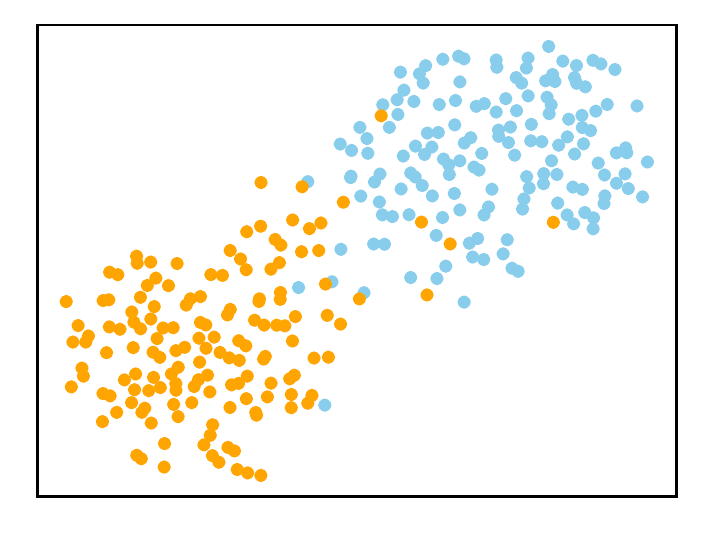}
}
\subfigure[\textsc{AdaVAE} with $\alpha=512$]{
\includegraphics[width=\tsnesize]{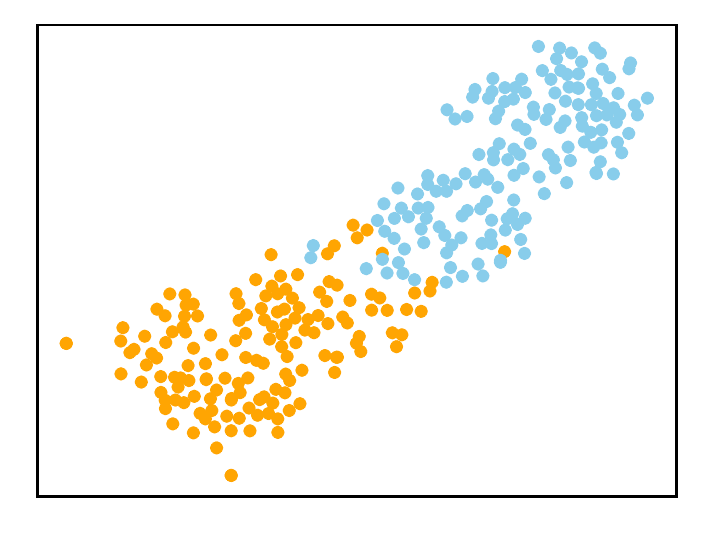}
}
\subfigure[\textsc{AdaVAE} with Fine-tuning]{
\includegraphics[width=\tsnesize]{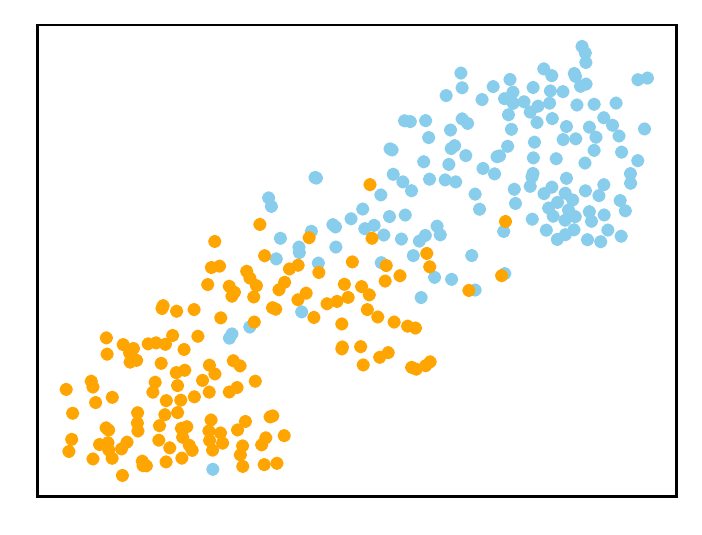}
}
\setlength{\abovecaptionskip}{0.1cm}
\caption{T-SNE plot for \textsc{AdaVAE} with different settings on the \texttt{YELP} dataset. The first two (a), (b) figures show \textsc{AdaVAE} with adapters of size $\alpha$. The last one (c) presents \textsc{AdaVAE} with finetuning method.}
\label{fig:tsne}
\end{figure*}

\begin{table*}
\small
\centering
\setlength\tabcolsep{1pt}
\begin{tabular}{ll} 
\toprule[1.5pt]
Source $S_A$                                                                                      & Target $S_B$                                                                                                                                      \\
\begin{tabular}[c]{@{}l@{}}$\bullet$ it is a very nice place to call home! \end{tabular}                                                           & \begin{tabular}[c]{@{}l@{}}$\bullet$ the food is so good and i seriously always feel like family.\end{tabular}                                           \\\hline
Input $S_C$                                                                                       & Output $S_D$                                                                                                                                      \\
\begin{tabular}[c]{@{}l@{}}$\bullet$ food was good, served in large portions.\end{tabular}             & \textcolor{blue}{\begin{tabular}[c]{@{}l@{}}\textcolor{black}{$\bullet$} food, and especially the appetizers, always exceeded \\\quad my expectations.\end{tabular}}                                   \\
\begin{tabular}[c]{@{}l@{}}$\bullet$ great experience every time at prestige animal clinic!\end{tabular} & \textcolor{blue}{\begin{tabular}[c]{@{}l@{}}\textcolor{black}{$\bullet$} best experience i have ever had in any restaurant, food \\\quad wise, the staff is amazing!\end{tabular}}                     \\
\begin{tabular}[c]{@{}l@{}}$\bullet$ i was very disappointed with the quality and service.\end{tabular}  & \textcolor{blue}{\begin{tabular}[c]{@{}l@{}}\textcolor{black}{$\bullet$} the food was absolutely horrible, i absolutely  never\\\quad tasted the food and the quality of the service.\end{tabular}}  \\
\bottomrule[1.5pt]
\end{tabular}
\caption{Sentence transfer via arithmetic $\boldsymbol{z_D} = \boldsymbol{z_B} - \boldsymbol{z_A} + \boldsymbol{z_C}$. Generated sentences are shown in \textcolor{blue}{blue}.}
\label{tab:analogy}
\end{table*}

The results are shown in Table~\ref{tab:controlgen}. We can draw the following conclusion: (1) $\textsc{AdaVAE}_{dec}$ with its decoder fully activated is much more competent than \textsc{AdaVAE} in all metrics, we believe this is because the guided text generation process depends more on decoder updating to produce sentences with different text labels. However, neither $\textsc{AdaVAE}_{dec}$ nor \textsc{AdaVAE} finetunes all parameters in the encoder, indicating our adaptive \textsc{GPT-2} encoder generates persist and robust $\mathcal{Z}$ space for further generation. (2) Compared with baseline models, $\textsc{AdaVAE}_{dec}$ can achieve the best performance in both G-score and BLEU-F1, which demonstrates that the proposed model is capable of generating controllable contexts that preserve human-like text features (high BLEU score). (3) However, there is still a big margin between our $\textsc{AdaVAE}_{dec}$ and baselines on Self-BLEU for text diversity. This is because the motivation to fully activate the decoder in the first place is to produce higher quality contents, we can access the model performance via the overall metric BLEU-F1 that takes this trade-off into consideration. 

We also present generated sentences with given sentiment in Table~\ref{tab:controlgen_sent}. For the negative label, there are words like ``terrible'' and ``not impressed'' representing negative sentiment. While for the positive label, words including ``great'', ``perfect'' and ``happy'' indicate the positive sentiment of generated sentences.
\subsection{Text Classification and Visualization via Latent Space $\mathcal{Z}$}
To validate that the proposed model is qualified as a textual feature extractor even with minimum trainable parameters. We conducted full-sized as well as low resource classification task. The percentage of activated parameter is \textbf{\#params.}. In Table~\ref{tab:cls}, \textbf{FB} means only the linear layer of a classifier was activated in training. All results were averaged on 5 runs with different random seeds. 

We view the model performances by the size of training corpus, (1) when the number of labeled training sample is very low (full \texttt{WNLI} / 10 or 100 training samples from $\texttt{YELP}_S$), \textsc{AdaVAE} can achieve better classification accuracy than fine-tuned \textsc{AdaVAE}, and is even superior than both fine-tuned and parameter-efficient \textsc{BERT} or \textsc{Optimus}. (2) For middle sized training data (full / 1,000$\sim$10,000 training samples from $\texttt{YELP}_S$), \textsc{AdaVAE} shows competitive performance compared with \textsc{BERT} and \textsc{Optimus} and generally better than fine-tuned \textsc{AdaVAE}. (3) For large-scale dataset (\texttt{SST-2}), the performance of \textsc{AdaVAE} is inferior than \textsc{BERT} and \textsc{Optimus} by around $6\%$. 

Though the performance of \textsc{AdaVAE} with adaptive components falls behind baselines on the large-scale dataset, it only activates very few parameters in the encoder to fulfill this task. These statistics demonstrate that \textsc{AdaVAE} with few activated parameters is competent to extract textual features like specialized PLM such as \textsc{BERT} or \textsc{Optimus}. We ascribe it to the structural modification of unmasked \textsc{GPT-2} transformers as the encoder of \textsc{AdaVAE} as well as \textit{Latent Attention}'s powerful knowledge learning ability from transformers. 

From the concluded results, as the increase of training data size, models with adaptive parameter-efficient components gradually loses their advantage on small-sized dataset compared to the fine-tuning models. This phenomenon is also reported by a concurrent work \cite{chen2022revisiting}, which verifies our observations. Since we did not change the adapter structure significantly, the training time of our model is $60\%$ to $100\%$ compared with fine-tuning \cite{ding2022delta}.

Further, we visualize the distribution of $\mathcal{Z}$ using T-SNE \cite{van2008visualizing} on $\texttt{YELP}_S$ in Figure~\ref{fig:tsne}. Firstly, the representations from \textsc{AdaVAE} with adaptive settings (Figure~\ref{fig:tsne} (a), (b)) can be better separated compared with fine-tuned one. Secondly, latent distribution with a bigger adapter size (Figure~\ref{fig:tsne} (b)) yields a more compact clustering in the figure but also means more activated training parameters. This demonstrates a trade-off between training parameters and representation learning ability of \textsc{AdaVAE}.

\begin{table}[!t]
\small
\centering
\setlength\tabcolsep{2pt}
\begin{tabular}{ll} 
\toprule[1.5pt]
0.0 & $\bullet$ the location is clean and the patio is great !                                                                                       \\
0.1 & \textcolor{blue}{\begin{tabular}[c]{@{}l@{}}\textcolor{black}{$\bullet$} the patio is in the middle of the block and the\\open right is better.\end{tabular}}                                                                \\
0.2 & \textcolor{blue}{\textcolor{black}{$\bullet$} the patio terrace is really nice and on the menu!}                                                                                   \\
0.3 & \textcolor{blue}{\begin{tabular}[c]{@{}l@{}}\textcolor{black}{$\bullet$} the kitchen is perfect, however the menu is small\\on the menu option. \end{tabular}}                                                                     \\
0.4 & \textcolor{blue}{\begin{tabular}[c]{@{}l@{}}\textcolor{black}{$\bullet$} after the reservation is open, the place is spacious\\and well organized. \end{tabular}}                                                                  \\
0.5 & \textcolor{blue}{\begin{tabular}[c]{@{}l@{}}\textcolor{black}{$\bullet$} the restaurant is spacious with plenty of room to\\choose from, even inside a typical yelp.\end{tabular}}  \\
0.6 & \textcolor{blue}{\begin{tabular}[c]{@{}l@{}}\textcolor{black}{$\bullet$} the menu is a perfect fit for a night stand, and the\\waiter's number is super friendly!\end{tabular}}    \\
0.7 & \textcolor{blue}{\begin{tabular}[c]{@{}l@{}}\textcolor{black}{$\bullet$} in addition to its extensive menu, the patio is\\absolutely a wonderful place!\end{tabular}}               \\
0.8 & \textcolor{blue}{\begin{tabular}[c]{@{}l@{}}\textcolor{black}{$\bullet$} if you are a fan of the service of a good\\restaurant, then definitely take a visit.\end{tabular}}        \\
0.9 & \textcolor{blue}{\textcolor{black}{$\bullet$} very attentive, especially for a non english tour.}                                                                                   \\
1.0 & $\bullet$ very special place and highly recommended .                                                                                          \\
\bottomrule[1.5pt]
\end{tabular}
\caption{Interpolating latent space $\boldsymbol{z_{\tau}} = \boldsymbol{z_1}\cdot (1 - \tau) + \boldsymbol{z_2}\cdot \tau$. Each row shows $\tau$, and generated sentence conditioned on $\boldsymbol{z_{\tau}}$ are shown in \textcolor{blue}{blue}.}
\label{tab:interpolation}
\end{table}

\subsection{Sentence Generation by Latent Manipulation}
We also conducted latent analogy and interpolation task. For a given sentence triplet $S_A, S_B, S_C$, the analogy task generates a sentence from source $S_A$ to target $S_B$ and with a similar style of $S_C$ as examples are shown in Table~\ref{tab:analogy}. We can tell from the table, that generated sentences absorb all given sentence features. For example, three output texts in the analogy task talk about food, which are relevant to Target $S_B$. When Input $S_C$ turns to negative, the Output $S_D$ also steers to the negative emotion. As for interpolation task, given a sentence pair $S_A, S_B$, latent interpolation generates texts with styles transfer from $S_A$ to $S_B$ by latent space traversal as examples are shown in Table~\ref{tab:interpolation}. From the interpolated example, generated texts mix the semantics and syntax of the two initial sentences (when $\tau$ is 0.0 or 1.0) and smoothly change from one to the other. 

\subsection{Ablation Study w.r.t. Transformer Layers}
We conducted experiments to select the best encoder\&decoder settings w.r.t. the number of layers in the transformer models. As shown in Table~\ref{tab:layer_ablation}, we varied both encoder layer number (denote as \textbf{Enc.}) and decoder layer number (denote as \textbf{Dec.}) from $[6, 8, 10, 12]$. And we find that: (1) models with fewer decoder layers show worse language modeling capacities, e.g., the model with 8 encoder layers and 12 decoder layers reaches the lowest \textbf{PPL} value as well as \textbf{-ELBO} value among other decoder settings. (2) \textsc{AdaVAE} with more encoder layers is not necessary, as the model with the best performance equips 8 encoder layers.

\begin{table}[!t]
\setlength\tabcolsep{2pt}
\centering
\small
\begin{tabular}{ccccccc}
\toprule[1.5pt]
\textbf{Enc.} & \textbf{Dec.} & \textbf{PPL} $\downarrow$ & \textbf{-ELBO} $\downarrow$ & \textbf{MI} $\uparrow$ & \textbf{AU} $\uparrow$ & \textbf{\#params.} $\downarrow$ \\\hline
6                   & 12                  & 16.53        & 122.62         & 7.50        & 32          & $14.34\%$          \\
8                   & 6                   & 28.02        & 142.46         & 7.52        & 32          & $10.36\%$          \\
8                   & 8                   & 22.42        & 135.15         & 7.51        & 32          & $13.23\%$          \\
8                   & 10                  & 21.74        & 129.40         & 7.46        & 32          & $14.04\%$          \\
8                   & 12                  & 15.49        & 125.56         & 7.55        & 32          & $14.66\%$          \\
10                  & 12                  & 15.68        & 122.52         & 7.66        & 32          & $14.98\%$          \\
12                  & 12                  & 16.45        & 122.64         & 7.39        & 32          & $15.30\%$          \\
\bottomrule[1.5pt]
\end{tabular}
\caption{Language modeling of \textsc{AdaVAE} with varied encoder/decoder layers.}
\label{tab:layer_ablation}
\end{table}
\section{Conclusion}
In this paper, we explored the first large-scale VAE system \textsc{AdaVAE} with unified parameter-efficient \textsc{GPT-2}s. \textsc{AdaVAE} is efficient to be trained because it freezes both PLM encoder\&decoder while adding trainable adapters for tasks. \textsc{AdaVAE} is elegant in construction, because it has unified encoder-decoder from \textsc{GPT-2} with the same word embedding space. \textsc{AdaVAE} is effective for language tasks, because experiments validate \textsc{AdaVAE} with proposed \textit{Latent Attention} has competent generative ability and potential feature extraction capacity. Tasks including language modeling, controllable text generation, low resource classification, and qualitative analysis confirm the superiority of the proposed model.

To explore the vastness and universality of \textsc{AdaVAE}, we plan to take more experiments w.r.t. conditional text generation, especially through the means of combining parameter-efficient method and conditional generation such as prompt tuning \cite{lester2021power}. As now more and more parameter-efficient methods, as well as more PLMs emerge, incorporating them into \textsc{AdaVAE} can be a meaningful extension for exploiting the potential of the proposed framework as efficient ``big VAEs''.


\bibliography{tacl2021}

\begin{thebibliography}{42}
\expandafter\ifx\csname natexlab\endcsname\relax\def\natexlab#1{#1}\fi

\bibitem[{Bowman et~al.(2015)Bowman, Vilnis, Vinyals, Dai, Jozefowicz, and
  Bengio}]{bowman2015generating}
Samuel~R Bowman, Luke Vilnis, Oriol Vinyals, Andrew~M Dai, Rafal Jozefowicz,
  and Samy Bengio. 2015.
\newblock Generating sentences from a continuous space.
\newblock \emph{arXiv preprint arXiv:1511.06349}.

\bibitem[{Chen et~al.(2022)Chen, Liu, Meng, and Liang}]{chen2022revisiting}
Guanzheng Chen, Fangyu Liu, Zaiqiao Meng, and Shangsong Liang. 2022.
\newblock Revisiting parameter-efficient tuning: Are we really there yet?
\newblock \emph{arXiv preprint arXiv:2202.07962}.

\bibitem[{Dai et~al.(2020)Dai, Gan, Cheng, Tao, Carin, and Liu}]{dai2020apo}
Shuyang Dai, Zhe Gan, Yu~Cheng, Chenyang Tao, Lawrence Carin, and Jingjing Liu.
  2020.
\newblock Apo-vae: Text generation in hyperbolic space.
\newblock \emph{arXiv preprint arXiv:2005.00054}.

\bibitem[{Devlin et~al.(2018)Devlin, Chang, Lee, and
  Toutanova}]{devlin2018bert}
Jacob Devlin, Ming-Wei Chang, Kenton Lee, and Kristina Toutanova. 2018.
\newblock Bert: Pre-training of deep bidirectional transformers for language
  understanding.
\newblock \emph{arXiv preprint arXiv:1810.04805}.

\bibitem[{Ding et~al.(2022)Ding, Qin, Yang, Wei, Yang, Su, Hu, Chen, Chan, Chen
  et~al.}]{ding2022delta}
Ning Ding, Yujia Qin, Guang Yang, Fuchao Wei, Zonghan Yang, Yusheng Su,
  Shengding Hu, Yulin Chen, Chi-Min Chan, Weize Chen, et~al. 2022.
\newblock Delta tuning: A comprehensive study of parameter efficient methods
  for pre-trained language models.
\newblock \emph{arXiv preprint arXiv:2203.06904}.

\bibitem[{Duan et~al.(2019)Duan, Xu, Pei, Han, and Li}]{duan2019pre}
Yu~Duan, Canwen Xu, Jiaxin Pei, Jialong Han, and Chenliang Li. 2019.
\newblock Pre-train and plug-in: Flexible conditional text generation with
  variational auto-encoders.
\newblock \emph{arXiv preprint arXiv:1911.03882}.

\bibitem[{Fang et~al.(2019)Fang, Li, Gao, Dong, and Chen}]{fang2019implicit}
Le~Fang, Chunyuan Li, Jianfeng Gao, Wen Dong, and Changyou Chen. 2019.
\newblock Implicit deep latent variable models for text generation.
\newblock \emph{arXiv preprint arXiv:1908.11527}.

\bibitem[{Fang et~al.(2021)Fang, Zeng, Liu, Bo, Dong, and
  Chen}]{fang2021transformer}
Le~Fang, Tao Zeng, Chaochun Liu, Liefeng Bo, Wen Dong, and Changyou Chen. 2021.
\newblock Transformer-based conditional variational autoencoder for
  controllable story generation.
\newblock \emph{arXiv preprint arXiv:2101.00828}.

\bibitem[{Fang et~al.(2022)Fang, Li, Shang, Jiang, Liu, and
  Yeung}]{fang2022controlled}
Xianghong Fang, Jian Li, Lifeng Shang, Xin Jiang, Qun Liu, and Dit-Yan Yeung.
  2022.
\newblock Controlled text generation using dictionary prior in variational
  autoencoders.
\newblock In \emph{Findings of the Association for Computational Linguistics:
  ACL 2022}, pages 97--111.

\bibitem[{Fu et~al.(2019)Fu, Li, Liu, Gao, Celikyilmaz, and
  Carin}]{fu2019cyclical}
Hao Fu, Chunyuan Li, Xiaodong Liu, Jianfeng Gao, Asli Celikyilmaz, and Lawrence
  Carin. 2019.
\newblock Cyclical annealing schedule: A simple approach to mitigating kl
  vanishing.
\newblock \emph{arXiv preprint arXiv:1903.10145}.

\bibitem[{Gururangan et~al.(2019)Gururangan, Dang, Card, and
  Smith}]{gururangan2019variational}
Suchin Gururangan, Tam Dang, Dallas Card, and Noah~A Smith. 2019.
\newblock Variational pretraining for semi-supervised text classification.
\newblock \emph{arXiv preprint arXiv:1906.02242}.

\bibitem[{He et~al.(2021)He, Zhou, Ma, Berg-Kirkpatrick, and
  Neubig}]{he2021towards}
Junxian He, Chunting Zhou, Xuezhe Ma, Taylor Berg-Kirkpatrick, and Graham
  Neubig. 2021.
\newblock Towards a unified view of parameter-efficient transfer learning.
\newblock \emph{arXiv preprint arXiv:2110.04366}.

\bibitem[{Hopfield(1982)}]{hopfield1982neural}
John~J Hopfield. 1982.
\newblock Neural networks and physical systems with emergent collective
  computational abilities.
\newblock \emph{Proceedings of the national academy of sciences},
  79(8):2554--2558.

\bibitem[{Houlsby et~al.(2019)Houlsby, Giurgiu, Jastrzebski, Morrone,
  De~Laroussilhe, Gesmundo, Attariyan, and Gelly}]{houlsby2019parameter}
Neil Houlsby, Andrei Giurgiu, Stanislaw Jastrzebski, Bruna Morrone, Quentin
  De~Laroussilhe, Andrea Gesmundo, Mona Attariyan, and Sylvain Gelly. 2019.
\newblock Parameter-efficient transfer learning for nlp.
\newblock In \emph{International Conference on Machine Learning}, pages
  2790--2799. PMLR.

\bibitem[{Hu et~al.(2021)Hu, Shen, Wallis, Allen-Zhu, Li, Wang, Wang, and
  Chen}]{hu2021lora}
Edward~J Hu, Yelong Shen, Phillip Wallis, Zeyuan Allen-Zhu, Yuanzhi Li, Shean
  Wang, Lu~Wang, and Weizhu Chen. 2021.
\newblock Lora: Low-rank adaptation of large language models.
\newblock \emph{arXiv preprint arXiv:2106.09685}.

\bibitem[{Hu et~al.(2017)Hu, Yang, Liang, Salakhutdinov, and
  Xing}]{hu2017toward}
Zhiting Hu, Zichao Yang, Xiaodan Liang, Ruslan Salakhutdinov, and Eric~P Xing.
  2017.
\newblock Toward controlled generation of text.
\newblock In \emph{International conference on machine learning}, pages
  1587--1596. PMLR.

\bibitem[{Lester et~al.(2021)Lester, Al-Rfou, and Constant}]{lester2021power}
Brian Lester, Rami Al-Rfou, and Noah Constant. 2021.
\newblock The power of scale for parameter-efficient prompt tuning.
\newblock \emph{arXiv preprint arXiv:2104.08691}.

\bibitem[{Lewis et~al.(2019)Lewis, Liu, Goyal, Ghazvininejad, Mohamed, Levy,
  Stoyanov, and Zettlemoyer}]{lewis2019bart}
Mike Lewis, Yinhan Liu, Naman Goyal, Marjan Ghazvininejad, Abdelrahman Mohamed,
  Omer Levy, Ves Stoyanov, and Luke Zettlemoyer. 2019.
\newblock Bart: Denoising sequence-to-sequence pre-training for natural
  language generation, translation, and comprehension.
\newblock \emph{arXiv preprint arXiv:1910.13461}.

\bibitem[{Li et~al.(2019)Li, He, Neubig, Berg-Kirkpatrick, and
  Yang}]{li2019surprisingly}
Bohan Li, Junxian He, Graham Neubig, Taylor Berg-Kirkpatrick, and Yiming Yang.
  2019.
\newblock A surprisingly effective fix for deep latent variable modeling of
  text.
\newblock \emph{arXiv preprint arXiv:1909.00868}.

\bibitem[{Li et~al.(2020)Li, Gao, Li, Peng, Li, Zhang, and Gao}]{li2020optimus}
Chunyuan Li, Xiang Gao, Yuan Li, Baolin Peng, Xiujun Li, Yizhe Zhang, and
  Jianfeng Gao. 2020.
\newblock Optimus: Organizing sentences via pre-trained modeling of a latent
  space.
\newblock \emph{arXiv preprint arXiv:2004.04092}.

\bibitem[{Li and Liang(2021)}]{li2021prefix}
Xiang~Lisa Li and Percy Liang. 2021.
\newblock Prefix-tuning: Optimizing continuous prompts for generation.
\newblock \emph{arXiv preprint arXiv:2101.00190}.

\bibitem[{Lucas et~al.(2019)Lucas, Tucker, Grosse, and
  Norouzi}]{lucas2019understanding}
James Lucas, George Tucker, Roger Grosse, and Mohammad Norouzi. 2019.
\newblock Understanding posterior collapse in generative latent variable
  models.

\bibitem[{Van~der Maaten and Hinton(2008)}]{van2008visualizing}
Laurens Van~der Maaten and Geoffrey Hinton. 2008.
\newblock Visualizing data using t-sne.
\newblock \emph{Journal of machine learning research}, 9(11).

\bibitem[{Mirza and Osindero(2014)}]{mirza2014conditional}
Mehdi Mirza and Simon Osindero. 2014.
\newblock Conditional generative adversarial nets.
\newblock \emph{arXiv preprint arXiv:1411.1784}.

\bibitem[{Park and Lee(2021)}]{park2021finetuning}
Seongmin Park and Jihwa Lee. 2021.
\newblock Finetuning pretrained transformers into variational autoencoders.
\newblock \emph{arXiv preprint arXiv:2108.02446}.

\bibitem[{Pelsmaeker and Aziz(2019)}]{pelsmaeker2019effective}
Tom Pelsmaeker and Wilker Aziz. 2019.
\newblock Effective estimation of deep generative language models.
\newblock \emph{arXiv preprint arXiv:1904.08194}.

\bibitem[{Pfeiffer et~al.(2020)Pfeiffer, Kamath, R{\"u}ckl{\'e}, Cho, and
  Gurevych}]{pfeiffer2020adapterfusion}
Jonas Pfeiffer, Aishwarya Kamath, Andreas R{\"u}ckl{\'e}, Kyunghyun Cho, and
  Iryna Gurevych. 2020.
\newblock Adapterfusion: Non-destructive task composition for transfer
  learning.
\newblock \emph{arXiv preprint arXiv:2005.00247}.

\bibitem[{Popel and Bojar(2018)}]{popel2018training}
Martin Popel and Ond{\v{r}}ej Bojar. 2018.
\newblock Training tips for the transformer model.
\newblock \emph{arXiv preprint arXiv:1804.00247}.

\bibitem[{Radford et~al.(2019)Radford, Wu, Child, Luan, Amodei, Sutskever
  et~al.}]{radford2019language}
Alec Radford, Jeffrey Wu, Rewon Child, David Luan, Dario Amodei, Ilya
  Sutskever, et~al. 2019.
\newblock Language models are unsupervised multitask learners.
\newblock \emph{OpenAI blog}, 1(8):9.

\bibitem[{Raffel et~al.(2019)Raffel, Shazeer, Roberts, Lee, Narang, Matena,
  Zhou, Li, and Liu}]{raffel2019exploring}
Colin Raffel, Noam Shazeer, Adam Roberts, Katherine Lee, Sharan Narang, Michael
  Matena, Yanqi Zhou, Wei Li, and Peter~J Liu. 2019.
\newblock Exploring the limits of transfer learning with a unified text-to-text
  transformer.
\newblock \emph{arXiv preprint arXiv:1910.10683}.

\bibitem[{Semeniuta et~al.(2017)Semeniuta, Severyn, and
  Barth}]{semeniuta2017hybrid}
Stanislau Semeniuta, Aliaksei Severyn, and Erhardt Barth. 2017.
\newblock A hybrid convolutional variational autoencoder for text generation.
\newblock \emph{arXiv preprint arXiv:1702.02390}.

\bibitem[{Shen et~al.(2020)Shen, Mueller, Barzilay, and
  Jaakkola}]{shen2020educating}
Tianxiao Shen, Jonas Mueller, Regina Barzilay, and Tommi Jaakkola. 2020.
\newblock Educating text autoencoders: Latent representation guidance via
  denoising.
\newblock In \emph{International Conference on Machine Learning}, pages
  8719--8729. PMLR.

\bibitem[{Subramanian et~al.(2018)Subramanian, Mudumba, Sordoni, Trischler,
  Courville, and Pal}]{subramanian2018towards}
Sandeep Subramanian, Sai~Rajeswar Mudumba, Alessandro Sordoni, Adam Trischler,
  Aaron~C Courville, and Chris Pal. 2018.
\newblock Towards text generation with adversarially learned neural outlines.
\newblock \emph{Advances in Neural Information Processing Systems}, 31.

\bibitem[{Vaswani et~al.(2017)Vaswani, Shazeer, Parmar, Uszkoreit, Jones,
  Gomez, Kaiser, and Polosukhin}]{vaswani2017attention}
Ashish Vaswani, Noam Shazeer, Niki Parmar, Jakob Uszkoreit, Llion Jones,
  Aidan~N Gomez, {\L}ukasz Kaiser, and Illia Polosukhin. 2017.
\newblock Attention is all you need.
\newblock \emph{Advances in neural information processing systems}, 30.

\bibitem[{Wang et~al.(2018)Wang, Singh, Michael, Hill, Levy, and
  Bowman}]{wang2018glue}
Alex Wang, Amanpreet Singh, Julian Michael, Felix Hill, Omer Levy, and Samuel
  Bowman. 2018.
\newblock Glue: A multi-task benchmark and analysis platform for natural
  language understanding.
\newblock In \emph{Proceedings of the 2018 EMNLP Workshop BlackboxNLP:
  Analyzing and Interpreting Neural Networks for NLP}, pages 353--355.

\bibitem[{Wang and Wan(2019)}]{wang2019t}
Tianming Wang and Xiaojun Wan. 2019.
\newblock T-cvae: Transformer-based conditioned variational autoencoder for
  story completion.
\newblock In \emph{IJCAI}, pages 5233--5239.

\bibitem[{Wang et~al.(2019)Wang, Gan, Xu, Zhang, Wang, Shen, Chen, and
  Carin}]{wang2019topic}
Wenlin Wang, Zhe Gan, Hongteng Xu, Ruiyi Zhang, Guoyin Wang, Dinghan Shen,
  Changyou Chen, and Lawrence Carin. 2019.
\newblock Topic-guided variational auto-encoder for text generation.
\newblock In \emph{NAACL-HLT (1)}.

\bibitem[{Xiao et~al.(2018)Xiao, Zhao, and Wang}]{xiao2018dirichlet}
Yijun Xiao, Tiancheng Zhao, and William~Yang Wang. 2018.
\newblock Dirichlet variational autoencoder for text modeling.
\newblock \emph{arXiv preprint arXiv:1811.00135}.

\bibitem[{Xu et~al.(2020)Xu, Cheung, and Cao}]{xu2020variational}
Peng Xu, Jackie Chi~Kit Cheung, and Yanshuai Cao. 2020.
\newblock On variational learning of controllable representations for text
  without supervision.
\newblock In \emph{International Conference on Machine Learning}, pages
  10534--10543. PMLR.

\bibitem[{Zhao et~al.(2018)Zhao, Kim, Zhang, Rush, and
  LeCun}]{zhao2018adversarially}
Junbo Zhao, Yoon Kim, Kelly Zhang, Alexander Rush, and Yann LeCun. 2018.
\newblock Adversarially regularized autoencoders.
\newblock In \emph{International conference on machine learning}, pages
  5902--5911. PMLR.

\bibitem[{Zhao et~al.(2017{\natexlab{a}})Zhao, Song, and
  Ermon}]{zhao2017infovae}
Shengjia Zhao, Jiaming Song, and Stefano Ermon. 2017{\natexlab{a}}.
\newblock Infovae: Information maximizing variational autoencoders.
\newblock \emph{arXiv preprint arXiv:1706.02262}.

\bibitem[{Zhao et~al.(2017{\natexlab{b}})Zhao, Zhao, and
  Eskenazi}]{zhao2017learning}
Tiancheng Zhao, Ran Zhao, and Maxine Eskenazi. 2017{\natexlab{b}}.
\newblock Learning discourse-level diversity for neural dialog models using
  conditional variational autoencoders.
\newblock \emph{arXiv preprint arXiv:1703.10960}.

\end{thebibliography}
\bibliographystyle{acl_natbib}
\newpage
\appendix
\section{Training Curves}
In order to show the effectiveness and stability of our training method, we plot the curves of KL weight, KL divergence, \textbf{PPL} scores, and \textbf{ELBO} scores of the \textsc{AdaVAE} model when training on the \texttt{Yelp} task as shown in Figure~\ref{fig:training_curve}. As the cyclic annealing of KL weight proceeds in training, KL divergence and \textbf{ELBO} values increase correspondingly, while the model \textbf{PPL} values decrease monotonously and show a convergent trending.
\newcommand{\mysize}{0.80\linewidth}
\begin{figure}[]
\centering
\subfigure[KL weight $\beta$ with 4 cyclic annealing.]{
\includegraphics[width=\mysize]{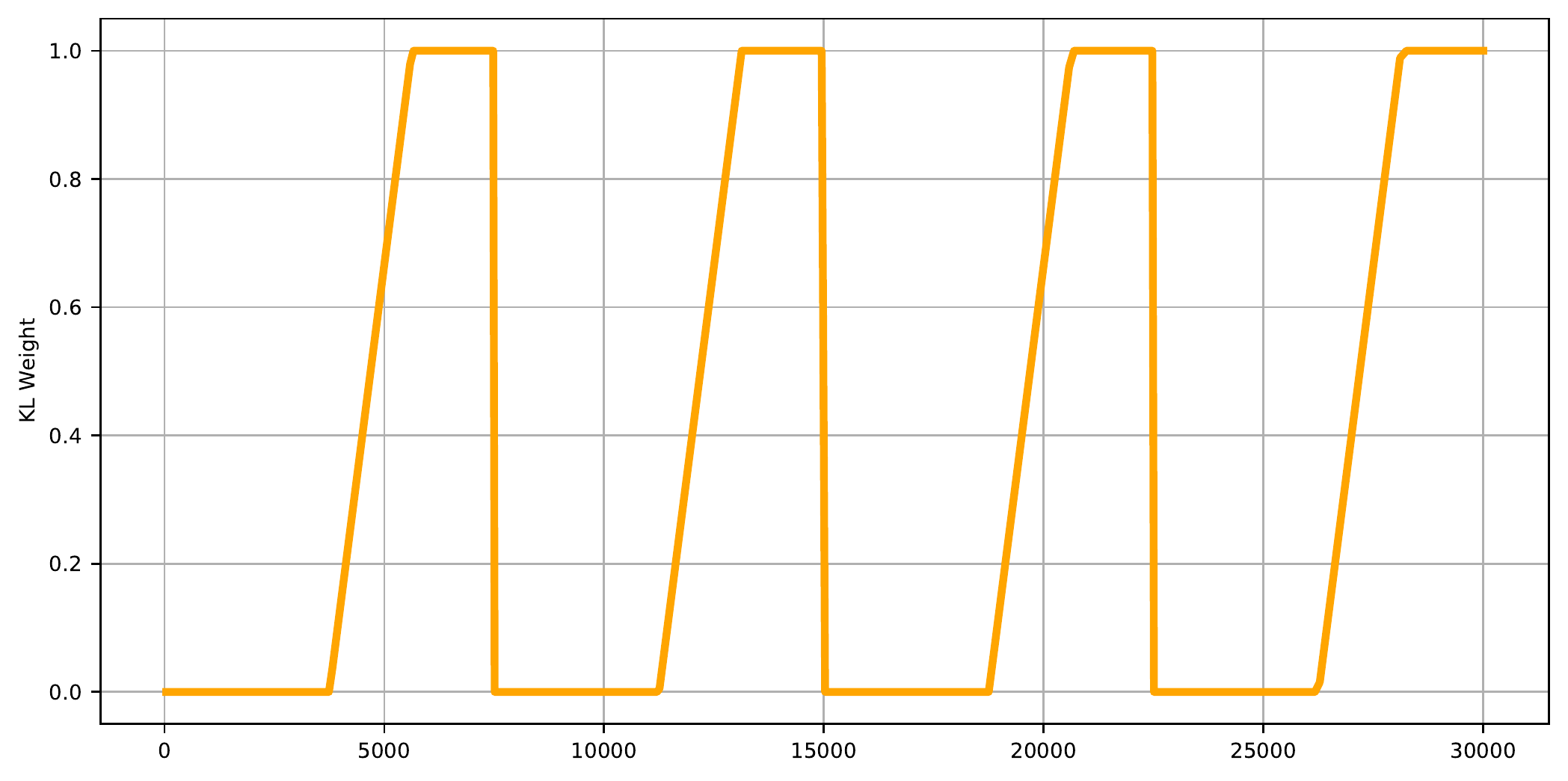}
}
\subfigure[KL divergence curve during training.]{
\includegraphics[width=\mysize]{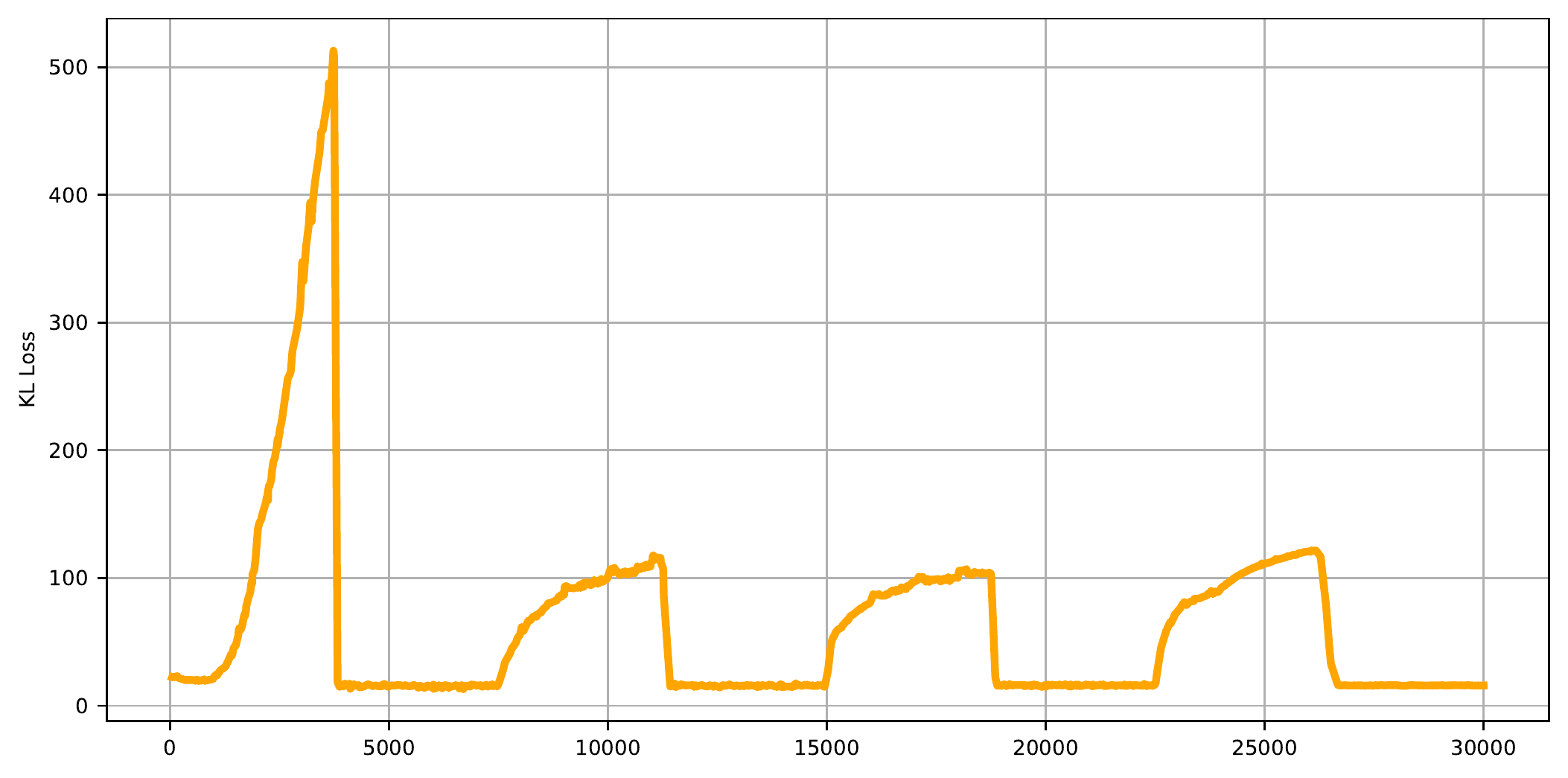}
}
\subfigure[Model \textbf{PPL} curve during training.]{
\includegraphics[width=\mysize]{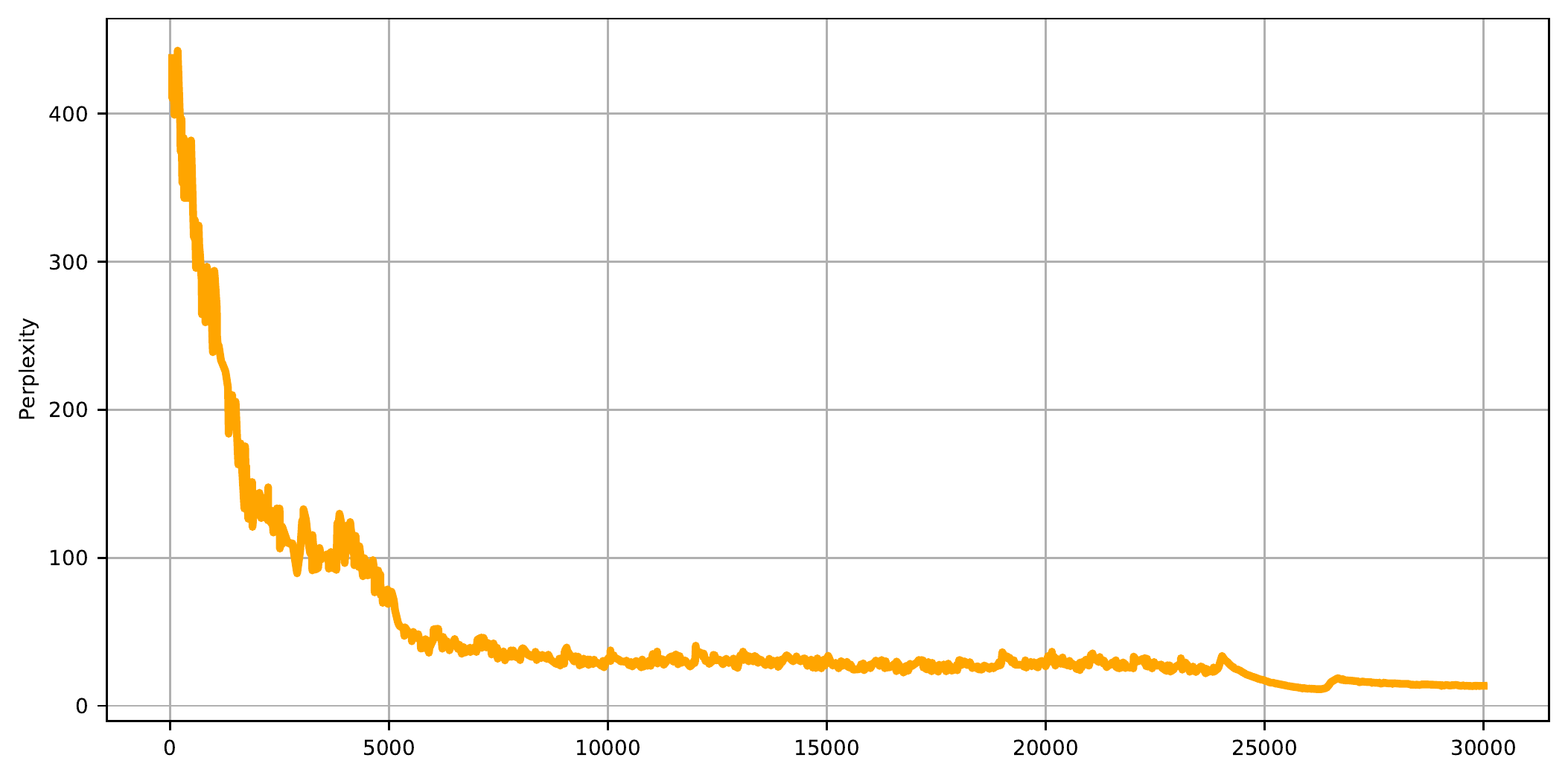}
}
\subfigure[\textbf{ELBO} loss curve during training.]{
\includegraphics[width=\mysize]{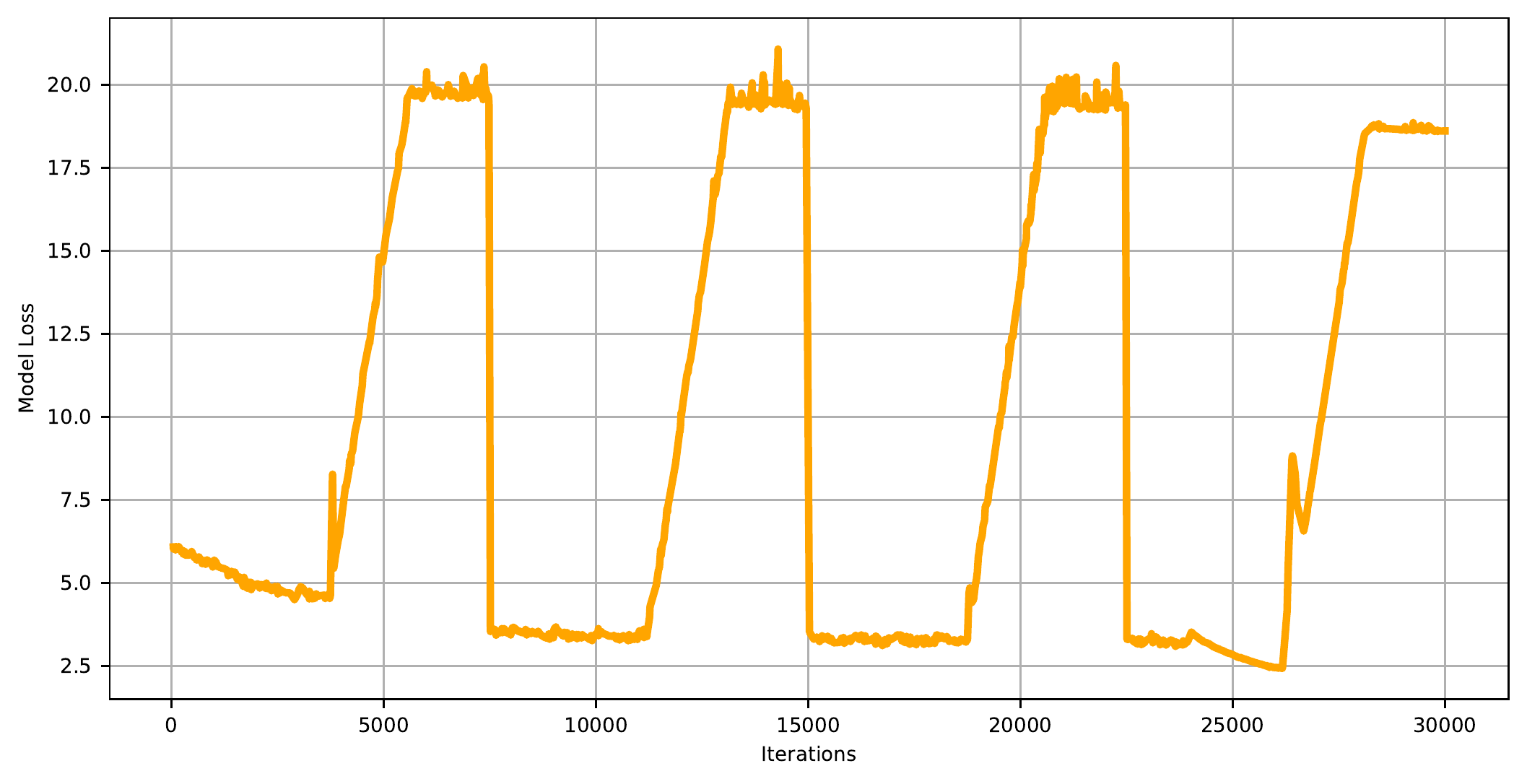}
}
\setlength{\abovecaptionskip}{0.1cm}
\caption{Training curve of \textsc{AdaVAE} w.r.t. different metrics on \texttt{Yelp} dataset.}
\label{fig:training_curve}
\end{figure}
\end{document}